\documentclass{article} %
\usepackage{iclr2027_conference,times}

\usepackage{amsmath}
\usepackage{amsfonts}
\usepackage{amssymb}
\usepackage{bm}
\usepackage{bbm}
\usepackage{stmaryrd}
\usepackage{multirow} 
\usepackage{todonotes} 
\newcommand{\hide}[1]{}

\usepackage{graphicx}
\usepackage{booktabs}
\usepackage[table]{xcolor}

\definecolor{oursrow}{gray}{0.93}
\newcommand{\ours}{\rowcolor{oursrow}}

\usepackage{comment}

\usepackage{amsthm}

\theoremstyle{plain}

\theoremstyle{definition}

\usepackage{url}
\usepackage{hyperref}

\newcommand{\quentin}[1]{\textcolor{black}{#1}}

\usepackage{amsmath,amsfonts,bm}
\usepackage{stmaryrd}

\def\eqref#1{equation~\ref{#1}}

\def\1{\bm{1}}

\def\vtheta{{\bm{\theta}}}

\def\vh{{\bm{h}}}

\def\vx{{\bm{x}}}
\def\vy{{\bm{y}}}

\def\vtheta{{\bm{\theta}}}

\def\vpsi{{\bm{\psi}}}

\def\vphi{{\bm{\phi}}}

\def\mD{{\bm{D}}}

\def\mI{{\bm{I}}}

\def\mK{{\bm{K}}}
\def\mL{{\bm{L}}}

\def\mS{{\bm{S}}}

\def\mW{{\bm{W}}}

\DeclareMathAlphabet{\mathsfit}{\encodingdefault}{\sfdefault}{m}{sl}
\SetMathAlphabet{\mathsfit}{bold}{\encodingdefault}{\sfdefault}{bx}{n}

\newcommand{\R}{\mathbb{R}}

\title{Semantic Uncertainty Quantification Needs Factual Equivalence}

\author{Joseph Hoche, Quentin Guimard, Gianni Franchi \\
AMIAD, P\^ole Recherche Palaiseau, France \\
\texttt{joseph.hoche@polytechnique.edu} \\
}

\iclrfinalcopy
\begin{document}

\maketitle

\begin{abstract}
Semantic uncertainty quantification for large language models rests on a common template: sample several answers, measure how much they agree, and treat disagreement as uncertainty.
We first formalize this template as two separate roles: an operator that compares two answers, and an aggregator that combines all pairwise comparisons into a scalar.
Existing methods differ almost entirely in how they aggregate, while taking the operator off the shelf, typically an NLI model or a generic sentence encoder.
\quentin{We show that this reliance on off-the-shelf operators is the primary bottleneck of semantic UQ: they do not accurately measure factual equivalence of multiple answers to the same question.}
\quentin{We resolve this with a deliberately simple recipe: a single encoder trained contrastively to isolate the targeted fact, utilizing synthetic data generated by an LLM and dataset both disjoint from all evaluation settings.}
\quentin{Integrating the resulting operator into existing methods improves performance on 120 of 126 evaluation settings (95\%) spanning 18 model--dataset combinations across language and vision-language models.}
\quentin{The best variant reaches 0.76 mean AUROC against 0.68 for the strongest baseline, while replacing the quadratic cross-encoder comparisons of entailment-based operators with one encoder pass per answer.}
The uniformity of the improvement supports the view that the operator, not the aggregator, is the limiting factor.
The same operator also improves single generation token-level estimators: the norm it assigns to each token measures how much that token bears on the answer, and reweighting token log-likelihoods accordingly sharpens the estimate.
\end{abstract}

\section{Introduction}

Despite their overwhelming popularity, Large Language Models (LLMs) and Large Vision-Language Models (LVLMs) are still prone to \emph{hallucinations}: they generate fluent, confident answers that are not grounded in their input~\citet{ji2023, liu2024survey}. Knowing when an output should not be relied upon has therefore become essential, which is the role of uncertainty quantification (UQ)~\citet{hernandez2015, gal2016dropout}.

\begin{figure}[t]
  \centering
  \includegraphics[width=\textwidth]{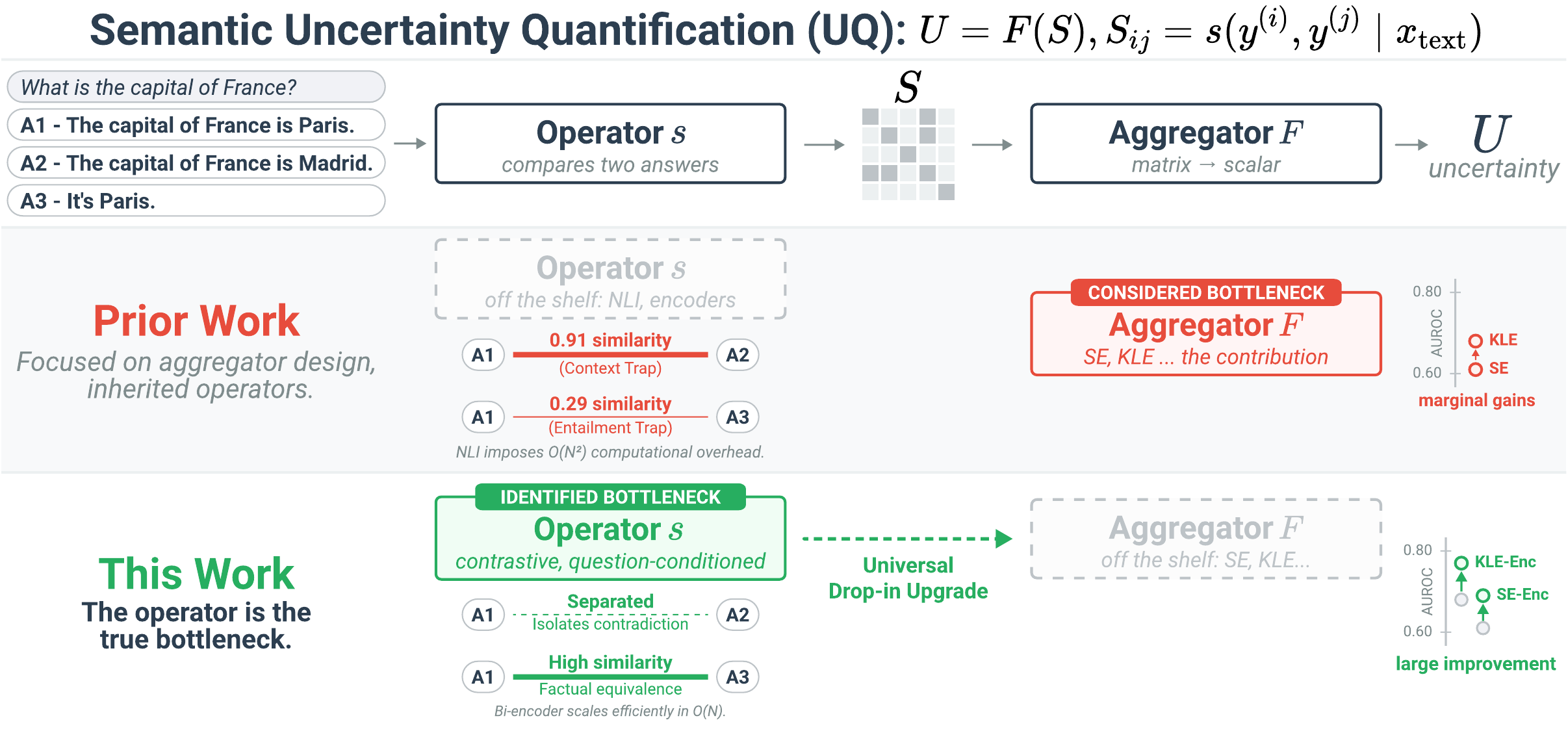}
  \caption{\textbf{The Operator Bottleneck in Semantic UQ.} Semantic uncertainty methods decompose into a pairwise comparison operator ($s$) and a structural aggregator ($F$). While prior work focuses on designing sophisticated aggregators, they inherit off-the-shelf operators that suffer from misaligned training objectives or severe scaling bottlenecks. We identify the operator as the true bottleneck and introduce a contrastively trained, question-conditioned bi-encoder. Serving as a universal drop-in replacement, our learned operator universally improves the performance of existing aggregators at a fraction of the computational cost of NLI baselines.}
  \label{fig:teasing}
\end{figure}

Early token-level methods operate directly on autoregressive token probabilities~\citet{malinin2021, aichberger2024rethinking}.
However, these measures conflate lexical variation with semantic variation. Given the question \textit{``What is the capital of France?''}, a model may output \textit{``The capital of France is Paris.''} (A1), \textit{``The capital of France is Madrid.''} (A2), or \textit{``It's Paris.''} (A3) (Fig.~\ref{fig:teasing}, top row).
A1 and A3 agree on the fact, yet receive different token-level scores because they differ in wording and structure.
A model that is entirely certain of the correct fact can thus appear highly uncertain simply due to lexical diversity.

Semantic UQ methods resolve this by evaluating generations in a space of meaning rather than vocabulary~\citet{farquhar2024detecting}.
As illustrated in Fig.~\ref{fig:teasing}, they share a common template: sample $N$ answers, compare them pairwise, and aggregate the resulting comparison structure into a scalar uncertainty score.
We formalize this template as $U = F(\mS)$, where an \emph{operator} $s$ scores pairwise semantic equivalence to form an $N \times N$ matrix $\mS$, and an \emph{aggregator} $F$ summarizes its structure.
Prior work has focused almost exclusively on designing sophisticated aggregators $F$: taking the entropy of semantic clusters (SE, CAE)~\citet{farquhar2024detecting}, computing von Neumann entropy over graph Laplacians (KLE)~\citet{nikitin2024kernel}, or analyzing the spectrum of a similarity matrix (COS)~\citet{chen2024}.
In every case, the operator $s$ is treated as a secondary design choice. Prior work reports that switching between off-the-shelf operators yields limited improvements~\citet{farquhar2024detecting}, which we argue occurs not because the choice is unimportant, but because none of these operators is structurally suited for the task.

This reliance on off-the-shelf operators is the primary bottleneck of semantic UQ.
As illustrated in the ``Prior Work'' row of Fig.~\ref{fig:teasing}, when existing operators evaluate generated text, they suffer from severe distribution shifts and misaligned objectives.
External sentence encoders evaluate general topical overlap, blinding them to targeted factual contradictions (e.g., scoring \textit{``The capital is Paris''} and \textit{``The capital is Madrid''} as highly similar).
Internal hidden states are optimized for autoregressive next-token prediction, which induces a strong recency bias that prioritizes shared syntactic suffixes over factual denotation.
Finally, Natural Language Inference (NLI) cross-encoders 
successfully incorporate the question but frame the comparison as strict directional entailment.
This fundamental misalignment causes them to artificially fracture meaning clusters when identical facts are expressed with natural, asymmetric verbosity.
Furthermore, evaluating entailment requires joint pairwise inference, imposing an $O(N^2)$ computational bottleneck.
No existing choice evaluates targeted factual equivalence conditioned on the question, aligns natively with the generative distribution of the LLM, and scales efficiently.

\quentin{In this work, we show that the operator, rather than the aggregator, is the true bottleneck of semantic UQ because off-the-shelf metrics fail to measure factual equivalence (Fig.~\ref{fig:teasing}, bottom row).}
Instead of designing a new aggregator, we treat the equivalence operator as a first-class object of study.
To bypass the flaws of human-annotated NLP datasets, we extract a native metric space from the language model itself. When an LLM samples multiple responses, it might produce both correct and incorrect answers.
This \emph{predictive variability} provides a source of weak supervision: correct answers collapse into a single semantic equivalence class, while incorrect answers diverge.

By utilizing this structural property, we introduce a question-conditioned bi-encoder trained via a contrastive triplet objective. Trained offline without any semantic annotations, \quentin{on a dataset disjoint from all evaluation settings,} our encoder learns to look through lexical overlap and isolate factual divergences. During inference, it produces a mathematically symmetric similarity matrix $\mS$ using only $O(N)$ forward passes, serving as a universal, drop-in replacement for any existing aggregator.

\textbf{Contributions.}
\textbf{(1)} We formalize semantic UQ as the composition of a comparison operator and an aggregator ($U = F(\mS)$), isolating the off-the-shelf operator as the primary bottleneck limiting current methods.
\textbf{(2)} We introduce a framework to extract a native semantic geometry using weak supervision from the LLM's own predictive variability, training a question-conditioned equivalence bi-encoder without manual semantic labels.
\textbf{(3)} We demonstrate that our operator acts as a universal upgrade: replacing the default operators in SE, CAE, KLE, and COS improves performance on %
\quentin{120 of 126 evaluation settings (95\%) spanning 18 model--dataset combinations with both LLMs and LVLMs,}
raising the mean AUROC from 0.68 to 0.76 while bypassing the $O(N^2)$ scaling bottleneck of entailment baselines.
\textbf{(4)} We show that the benefits extend beyond semantic aggregation: the geometric norm of our bi-encoder assigns an intrinsic importance score to individual tokens, yielding a lightweight reweighting mechanism that strictly improves single-generation token-level methods.

\section{Related Work}
\label{sec:related_work}

\textbf{Non-semantic Uncertainty Quantification in LLMs and LVLMs.}
{Non-semantic UQ methods estimate uncertainty without explicitly comparing
the meaning of several generations.
Some approaches ask the model to report its own confidence, directly or in
natural language~\citet{kadavath2022,lin2022teaching}.
Others use information from a single generation, such as token
probabilities, likelihood-based scores, or internal model
representations~\citet{malinin2021,murray2018,ren2023,azaria2023the,li2023}.
Another line of work perturbs the input or the decoding process and measures
how much the prediction changes~\citet{gao2024,zhang2025,li2026semantic}.
}

\textbf{Semantic Uncertainty Quantification in LLMs and LVLMs.}
{The closest works to ours sample several responses and measure their agreement
in semantic space~\citet{kuhnsemantic,farquhar2024detecting,lin2024generating}.
Recent methods improve this idea by using local semantic density~\citet{qiu2024semantic}, pairwise semantic similarities~\citet{nguyen-etal-2025-beyond}, semantic embeddings~\citet{grewal2024improving}, geometric representations of uncertainty~\citet{li2026semantic}, or Gaussian-process models for LVLMs~\citet{hoche2025improving}.
Other work studies how to reduce the sampling cost of semantic UQ~\citet{park2026efficient}, while recent analyses stress that semantic
disagreement should not be confused directly with reliability~\citet{hoche2026position}.
Our work is complementary: rather than proposing a new way to aggregate
semantic disagreement, we focus on the comparison operator itself.
Existing methods typically rely on NLI models, internal LLM features, or
general-purpose sentence embeddings, whereas we learn a dedicated
question-conditioned semantic encoder directly from LLM generations.
}

\section{The Operator Bottleneck in Semantic UQ}
\label{sec:bottleneck}

Semantic UQ methods estimate uncertainty by measuring agreement across multiple model generations rather than relying on token-level probabilities. While the literature presents a diverse array of estimators, we observe that they all share a common structural decomposition.

\subsection{Formalizing Semantic UQ: an Operator Followed by an Aggregator}
\label{sec:formalism}

Given a textual query $\vx_{\mathrm{text}}$ and an autoregressive language model $p_{\vtheta}$, we sample $N$ responses $\vy^{(i)} \sim p_{\vtheta}(\cdot \mid \vx_{\mathrm{text}})$. Every semantic estimator proceeds in two stages. First, a pairwise comparison \emph{operator} $s$ evaluates how far two generations express the same answer, populating a comparison matrix $\mS \in \mathbb{R}^{N \times N}$:
\begin{equation}
\mS_{ij} = s\bigl(\vy^{(i)}, \vy^{(j)} \mid \vx_{\mathrm{text}}\bigr).
\label{eq:operator_S}
\end{equation}
Second, an \emph{aggregator} $F$ summarizes this matrix into an uncertainty score $U = F(\mS)$.

Prior work has focused predominantly on the design of the aggregator $F$. Semantic Entropy (SE) and Cluster-Assignment Entropy (CAE)~\citet{kuhnsemantic, farquhar2024detecting} reduce $\mS$ to a discrete partition and compute the Shannon entropy over the resulting clusters. Kernel Language Entropy (KLE)~\citet{nikitin2024kernel} treats $\mS$ as an adjacency matrix and computes the von Neumann entropy of its graph Laplacian. COS~\citet{chen2024} computes the spectral entropy of $\mS$ directly.

Conversely, the operator $s$ is treated as a secondary design choice, inherited off-the-shelf from components built for other tasks. SE, CAE, and KLE rely on Natural Language Inference (NLI) cross-encoders trained to predict directional entailment on human-written datasets (MNLI). %

COS relies on cosine similarities over either internal autoregressive hidden states (COS-int) or external general-purpose sentence encoders (COS-ext). 
A detailed formalization of these comparison matrices and their integration into existing baselines is provided in App.~\ref{supp:formalism}.

\subsection{Diagnostic Benchmark: How Existing Operators Fail}
\label{sec:diagnostics}

This design choice also creates a computational bottleneck: NLI cross-encoders compare
answers pairwise, requiring $O(N^2)$ expensive forward passes to build $\mS$.
More importantly, these borrowed representations are not designed for the
predictive variability of generative models. We therefore build a diagnostic
benchmark to study the main off-the-shelf operators.
Each experiment below is built from 1{,}000 pairs drawn from sampled and greedy
generations of the same pipeline as our training data (Sec.~\ref{sec:method}); reported
scores are averages over these pairs (details in App.~\ref{supp:limit_op}).

{\textbf{The Context Trap (COS-ext).}
External sentence encoders mainly capture overall semantic similarity and do
not directly use the question.
We compare a correct answer with a rewritten
version where only the key fact is changed.
Although the two answers contradict each other, 
external encoder gives a very high mean
cosine similarity of $0.91$.
This shows that lexical and topical overlap can hide the factual difference that
UQ should detect.
}

{
\textbf{The Autoregressive Bias (COS-int).}
Internal LLM representations are optimized for next-token prediction rather
than answer equivalence.
For contradictory pairs with a shared suffix, such as
\textit{``Paris is the capital.''} and
\textit{``Madrid is the capital.''}, 
internal representations
give a high mean similarity of
$0.87$.
Hence, the shared local context can dominate the representation and hide the
factual disagreement that should be detected by UQ measures.
}

{\textbf{The Entailment Trap (NLI Cross-Encoders\hide{\& LLMs}).}
NLI models are trained for directional entailment, not for deciding whether
two answers express the same fact.
To illustrate this, we compare pairs of correct generations with very
different lengths, such as \textit{``It is Paris.''} and
\textit{``The capital city of France is indeed Paris.''}.
Although both answers mean the same thing, NLI cross-encoders give a low mean
similarity of $0.29$.
This shows that entailment can split semantically equivalent answers simply
because they differ in wording or verbosity.}

\textbf{The Distribution Mismatch (OOD Syntax).} Beyond architectural flaws, supervised NLI metric spaces suffer from a fundamental domain shift. Operators trained on human-annotated text (MNLI) expect well-formed, grammatically complete sentences. To quantify this, we isolate 1,000 pairs comparing the LLM's greedy generated correct sentence (e.g., \textit{``The capital of France is Paris.''}) against the dataset's raw ground-truth entity (e.g., \textit{``Paris''}). {NLI cross-encoders give a low mean similarity of only $0.27$.
This shows that NLI can confuse differences in form with differences in
meaning, motivating an operator trained directly on LLM generations.}%

\subsection{Desiderata for an Ideal Operator}
\label{sec:desiderata}

These diagnostic failures dictate the exact requirements for a semantic UQ operator. We define an ideal operator $s^\star$ as one that satisfies three core desiderata:

\noindent\textbf{(D1) Invariance to Generative Syntax:} To resolve the COS-int autoregressive bias and the NLI distribution mismatch, the metric space must be robust to the specific distributional artifacts of $p_{\vtheta}$. It must evaluate semantic meaning regardless of asymmetric verbosity, single-word fragments, or shared causal suffixes.

\noindent\textbf{(D2) Question-Conditioned Factual Equivalence:} To resolve the COS-ext context trap and the NLI entailment trap, $s^\star$ must be explicitly anchored to $\vx_{\mathrm{text}}$. It must not measure general topical similarity or directional logic, but specifically evaluate whether two generations denote the exact same factual answer to the given query.

\noindent\textbf{(D3) Computational Scalability:} The operator must isolate sequence representations independently to construct the similarity matrix in $O(N)$ transformer passes. Relying on cross-encoders
to evaluate every pair incurs an $O(N^2)$ bottleneck that renders UQ prohibitively expensive at scale.

Satisfying these desiderata requires abandoning off-the-shelf metric spaces and external NLP datasets. In Section~\ref{sec:method}, we demonstrate how to extract an operator fulfilling all these properties by leveraging the LLM's own generative distribution.

\section{Learning Semantic Geometry from Predictive Variability}
\label{sec:method}

\begin{figure}[t]
  \centering
  \includegraphics[width=\textwidth]{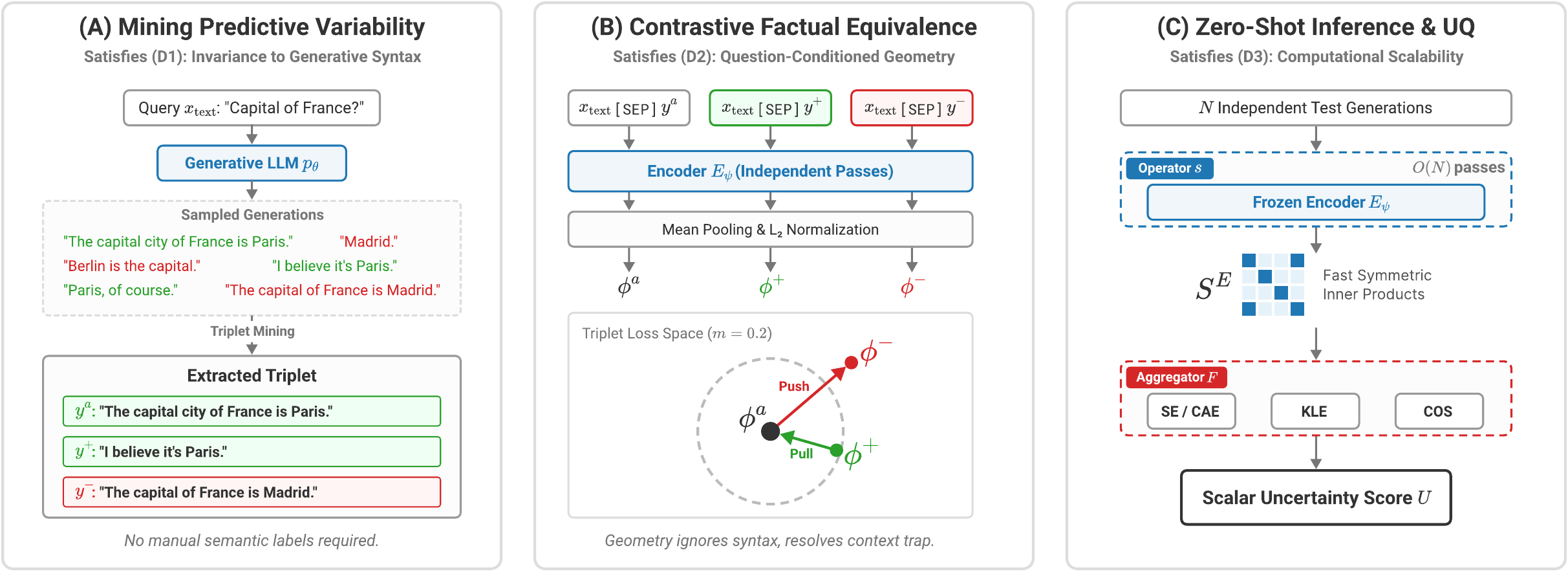}
  \caption{Learning Semantic Geometry from Predictive Variability. \textbf{(A) Mining Predictive Variability (Section~\ref{sec:weak_supervision}):} We extract positive and negative semantic triplets by evaluating repeated LLM generations for factual correctness, naturally capturing the model's native syntax distribution. \textbf{(B) Contrastive Factual Equivalence (Sections~\ref{sec:architecture}~and~\ref{sec:contrastive_objective}):} A shared bi-encoder contextualizes the answers against the query. The contrastive margin separates hard negatives (incorrect answers to the same question) from valid paraphrases. \textbf{(C) Zero-Shot Inference \& UQ (Section~\ref{sec:integration}):} The frozen encoder generates symmetric $O(N)$ comparison matrices that drop seamlessly into existing aggregators, while its geometric norm enables zero-cost token reweighting.}
  \label{fig:method}
\end{figure}

To satisfy the desiderata derived in Section~\ref{sec:bottleneck}, we introduce a framework that trains a comparison operator directly on the relation it is meant to express.
We extract a native semantic geometry by exploiting the predictive variability of the target language model $p_{\vtheta}$ (Fig.~\ref{fig:method}).
Specifically, we initialize a pre-trained bi-encoder (Section~\ref{sec:architecture}), construct a weakly supervised training dataset of semantic triplets by mining the LLM's own generations (Section~\ref{sec:weak_supervision}), and fine-tune the encoder using a contrastive objective to isolate factual equivalence (Section~\ref{sec:contrastive_objective}).

\subsection{The Architecture: Question-Conditioned Bi-Encoding}
\label{sec:architecture}

We begin by defining the trainable module and its inference architecture. To %
{comply with} \textbf{(D3)} (Computational Scalability), we avoid the $O(N^2)$ cross-attention bottleneck by adopting a bi-encoder framework. Our method is agnostic to the specific base model; we initialize a pre-trained transformer encoder $E_{\vpsi}$ (e.g., DeBERTa~\citet{he2020deberta}), which we will fine-tune to independently contextualize each generation against the query. 

Specifically, $E_{\vpsi}$ takes the question $\vx_{\mathrm{text}}$ and a single generated response $\vy^{(i)}$, joined by a separator token, and maps the sequence to a normalized dense vector $\bm{\phi}_i$:
\begin{equation}
  \bm{\phi}_i = \frac{\bar{\vh}_i}{\|\bar{\vh}_i\|_2}, \qquad \bar{\vh}_i = \frac{1}{T'} \sum_{t=1}^{T'} \vh_t \in \mathbb{R}^d,
  \label{eq:encoder}
\end{equation}
where $\vh_t$ are the contextualized hidden states of the final layer over $E_{\vpsi}\bigl(\vx_{\mathrm{text}} \,\texttt{[SEP]}\, \vy^{(i)}\bigr)$ and $T'$ is the length of the concatenated string. 

The semantic operator $s_{\vpsi}$ is defined as the inner product of two such embeddings:
\begin{equation}
s_{\vpsi}\bigl(\vy^{(i)}, \vy^{(j)} \mid \vx_{\mathrm{text}}\bigr) = \langle \bm{\phi}_i, \bm{\phi}_j \rangle.
\label{eq:operator_cosine}
\end{equation}

By keeping the question in the context window, this structural choice fulfills the first requirement of \textbf{(D2)} (Question-Conditioning), anchoring the comparison to the specific point at issue. Furthermore, because the transformer evaluates each generation only once, populating the symmetric matrix $\mS$ requires exactly $N$ forward passes followed by fast vector dot-products, reducing the operational cost from $O(N^2)$ to $O(N)$.

\subsection{The Training Data: Weak Supervision via Predictive Variability}
\label{sec:weak_supervision}

Satisfying \textbf{(D1)} (Invariance to Generative Syntax) requires $E_{\vpsi}$ to be robust to the specific conversational artifacts, varying verbosity, and single-word fragments native to the generative distribution. However, obtaining manual pairwise semantic annotations for raw LLM generations would be prohibitively expensive and defeat the purpose of an automated metric. 

Instead, we exploit a structural property of language models: when a query is sampled repeatedly from $p_{\vtheta}$, the model produces a mix of correct and incorrect answers. We demonstrate that this predictive variability provides a natural source of weak supervision (Fig.~\ref{fig:method}A). Let $z^\star$ denote the semantic equivalence class of the correct factual answer, and let $g_{\vx}(\vy)$ denote the underlying semantic class of a generated response. We define a binary correctness variable $C_i$:
\begin{equation}
  C_i = \mathbbm{1} \bigl[ g_{\vx}(\vy^{(i)}) = z^\star \bigr].
\end{equation}
We rely on an %
\textbf{assumption}: all correct responses to the same query, regardless of their length or phrasing, belong to the same semantic equivalence class. Therefore, correctness naturally induces exact pairwise semantic relations whenever at least one generation is correct:
\begin{align}
  C_i = C_j = 1 \quad &\Longrightarrow \quad g_{\vx}(\vy^{(i)}) = g_{\vx}(\vy^{(j)}), \\
  C_i \neq C_j \quad &\Longrightarrow \quad g_{\vx}(\vy^{(i)}) \neq g_{\vx}(\vy^{(j)}).
\end{align}

This formulation never requires an explicit supervision signal of the form \textit{``$\vy^{(i)}$ and $\vy^{(j)}$ mean the same thing.''} Instead, mining the outputs of $p_{\vtheta}$ automatically yields positive semantic pairs ($\vy_k^{a}, \vy_k^{+}$) comprising two correct generations, and negative semantic pairs ($\vy_k^{a}, \vy_k^{-}$) comprising one correct and one incorrect generation. Because these pairs are drawn directly from $p_{\vtheta}$, the metric space natively learns to map across the exact syntax and verbosity distributions encountered during inference.

To construct the training data, we sample multiple responses from \texttt{Llama-3.1-8B} on the NQ-OPEN dataset. An LLM-as-a-judge (\texttt{Qwen2.5-32B-Instruct}) independently labels each generation as correct or incorrect against the reference text (see App.~\ref{supp:llm_judge}). 

\subsection{The Objective: Contrastive Factual Equivalence}
\label{sec:contrastive_objective}

While the architecture provides the necessary question-conditioning, fulfilling the remainder of \textbf{(D2)} (Factual Equivalence) requires explicitly optimizing $E_{\vpsi}$ to measure targeted factual agreement rather than general topical similarity. We fine-tune the base encoder to organize the embedding space strictly according to the correctness-induced triplets (Fig.~\ref{fig:method}B).

For each triplet index $k$ in a batch of size $B$, containing an anchor, a positive, and a negative text sequence, we extract their respective normalized embeddings $(\bm{\phi}_k^{a}, \bm{\phi}_k^{+}, \bm{\phi}_k^{-})$ using Eq.~\ref{eq:encoder}, and minimize the contrastive triplet loss:
\begin{equation}
  \mathcal{L}_{\mathrm{triplet}} = \frac{1}{B} \sum_{k=1}^{B} \Bigl[
  \bigl\langle \bm{\phi}_k^{a}, \bm{\phi}_k^{-} \bigr\rangle -
  \bigl\langle \bm{\phi}_k^{a}, \bm{\phi}_k^{+} \bigr\rangle + m
  \Bigr]_{+},
  \label{eq:triplet_loss}
\end{equation}
where $m = 0.2$ is the margin. In high-dimensional spaces, unrelated text representations naturally concentrate near orthogonality ($\langle \bm{\phi}_k^a, \bm{\phi}_k^- \rangle \approx 0$). A margin of $m=0.2$ provides sufficient geometric separation to isolate factual contradictions, while remaining highly permissive of valid syntactic variations and varying verbosity within the positive pairs.

This triplet structure inherently defines a distribution of \emph{hard negatives}. Because the negative sequence $\vy_k^{-}$ is sampled as an incorrect response to the exact same question as the anchor $\vy_k^{a}$, it shares the overall topic and frequently mirrors the syntactic structure, differing only in the tokens conveying the factual claim. Separating these representations cannot be achieved through general topical similarity; the encoder is forced to place its geometric weight exclusively on the specific factual denotation answering the query, perfectly satisfying \textbf{(D2)}.

Crucially, this pipeline requires absolutely no semantic equivalence annotations or UQ confidence labels. While the training dataloader leverages binary correctness labels to construct triplets, the geometric objective forces $E_{\vpsi}$ to learn a generalized function of \emph{equivalence}, not absolute truth. Consequently, at inference time, the frozen operator $s_{\vpsi}$ successfully groups generations that agree, regardless of whether they agree on the correct answer. As detailed in App.~\ref{supp:wrong_answer_clustering}, it successfully clusters sets of identical factual hallucinations, which is exactly the structure required for downstream semantic UQ aggregators.

\section{A Universal Drop-In for Semantic UQ}
\label{sec:integration}

Because the semantic UQ template $U = F(\mS)$ decouples the metric space from the structural summary, the operator learned in Section~\ref{sec:method} can be substituted into existing estimators without modifying their aggregators.
We denote the symmetric comparison matrix produced by our operator as $\mS^{\mathrm{E}}$, where $\mS^{\mathrm{E}}_{ij} = \langle \bm{\phi}_i, \bm{\phi}_j \rangle$, and mark each upgraded variant with the subscript $\mathrm{Enc}$.

In every case, the original aggregator $F$ is applied exactly as published. The only change is the matrix it receives, isolating the performance gains entirely to the resolution of the operator bottleneck.

\subsection{Upgrading Semantic Aggregators}

Natural language inference is fundamentally directional, meaning existing semantic aggregators must enforce bidirectional entailment post-hoc to approximate equivalence. By generating symmetric inner-product matrices directly, our operator eliminates redundant directed evaluations and guarantees that $\mS^{\mathrm{E}}$ is a valid positive semidefinite Gram matrix, natively satisfying the spectral assumptions required by downstream graph and eigenvalue methods.

\textbf{Semantic Entropy (SE) and Cluster-Assignment Entropy (CAE)} both consume a binary adjacency matrix to partition generations into meaning clusters. We replace their off-the-shelf NLI matrix by thresholding $\mS^{\mathrm{E}}$: two generations are assigned to the same cluster when $\mS^{\mathrm{E}}_{ij} > \tau$, with a fixed threshold $\tau = 0.5$. The resulting partition is fed to the original entropy aggregators unchanged, yielding $\mathrm{SE}_{\mathrm{Enc}}$ and $\mathrm{CAE}_{\mathrm{Enc}}$.

\textbf{Kernel Language Entropy (KLE)} consumes a continuous similarity matrix, traditionally built from soft NLI entailment probabilities. We pass the continuous matrix $\mS^{\mathrm{E}}$ in its place and apply the identical graph Laplacian, heat/Mat\'ern kernel diffusion, and von Neumann entropy computations, yielding $\mathrm{KLE\text{-}Heat}_{\mathrm{Enc}}$ and $\mathrm{KLE\text{-}Matern}_{\mathrm{Enc}}$.

\textbf{Eigenvalue-based scores (COS)} already operate on cosine similarities, previously derived from question-agnostic external encoders or prefix-biased internal hidden states. We replace those flawed representations with our question-conditioned embeddings $\bm{\phi}_i$ and apply the original spectral entropy computation to yield $\mathrm{COS}_{\mathrm{Enc}}$.

\subsection{Lightweight Token-Level Reweighting}
\label{sec:token_level}

The geometric space the operator constructs also carries information at a finer
granularity, from which a measure of token importance can be read without any additional
training.

Writing each contextual representation as a magnitude and a direction,
the pooling of Eq.~\ref{eq:encoder} becomes a
weighted sum of unit directions, $\bm{\phi} \propto \sum_t \|\vh_t\|_2 \, \bar{\vh}_t$,
in which each token is weighted by its norm. Since the operator compares generations
solely through the normalized vector $\bm{\phi}$, only the directions carry meaning, and
the norms determine how much each token steers that direction. This allocation is a free
byproduct of the triplet loss: separating an incorrect answer from a correct one requires
placing norm on the tokens that carry the factual claim and withholding it from those that
vary freely across paraphrases. We therefore read $\|\vh_t\|_2$ as a measure of semantic
importance.

This upgrades the uniform averaging of token-level entropy (TE). Normalizing the norms of
the generation's tokens into weights over those positions, discarding the query and
special tokens, gives
\begin{equation}
w_t = \frac{\|\vh_t\|_2}{\sum_{t'=1}^{T} \|\vh_{t'}\|_2},
\qquad
U_{\mathrm{TE}_{\mathrm{Enc}}}(\vy \mid \vx, \vtheta) = \sum_{t=1}^{T} w_t \,
H_t(\vx, \vy_{<t}, \vtheta),
\qquad
{{\footnotesize\text{with } H_t \text{\footnotesize\ as in Eq.~(\ref{eq:token_level_entropy})}}}
\label{eq:te_enc}
\end{equation}
Like TE, this requires a single generation, but down-weights structurally necessary tokens
that carry no answer content. Because the encoder and the generating model use different
tokenizers, we apply a sequence alignment to map the norms onto the generative
vocabulary before computing the sum.

\section{Experiments}
\label{sec:experiments}

\subsection{Setup}
\textbf{Models and Datasets.} We evaluate on six models spanning both modalities. For text-only question answering we
use \texttt{Qwen2.5-7B}~\citet{bai2025qwen2}, \texttt{Phi-3.5-mini}~\citet{abdin2024phi} and
\texttt{Mistral-7B}~\citet{Jiang2023Mistral7}; for visual question answering we use
\texttt{idefics2-8B}~\citet{laurenccon2023obelics}, \texttt{Qwen2.5-VL-7B}~\citet{bai2025qwen2} and
\texttt{llava-1.5-7B}~\citet{liu2024improved}. None of these models is involved in the construction of
the training set of Sec.~\ref{sec:method}, which is built from generations of
\texttt{Llama-3.1-8B} alone. We evaluate on six datasets, again covering both settings. The visual question answering
benchmarks are \textbf{ADVQA}~\citet{li2021adversarial}, \textbf{OKVQA}~\citet{marino2019ok} and
\textbf{VizWiz}~\citet{gurari2018vizwiz}; the textual ones are \textbf{HotpotQA}~\citet{yang2018hotpotqa},
\textbf{TriviaQA}~\citet{joshi2017triviaqa} and \textbf{WebQuestions}~\citet{berant-etal-2013-semantic}. Each model is evaluated on
the three datasets of its modality, giving the 18 model--dataset combinations reported
throughout. The exact splits can be found in App.~\ref{supp:eval_details}. \textbf{NQ-OPEN}~\citet{lee2019latent}, used to train the operator, appears in none of them. 

\textbf{Generation.} For every question we sample $N = 20$ answers with temperature $1.0$, top-$p=0.9$, top-$k=50$, and additionally record the greedy generation together with its token-level
distributions, which the token-level estimators require. Additional details can be found in App.~\ref{supp:eval_details}.

\textbf{Correctness labels.} Following prior work~\citet{farquhar2024detecting}, correctness is assessed on the greedy generation: an
LLM-as-a-judge, here \texttt{Llama-3.1-70B-Instruct}, compares it to the reference answer and
returns a binary label, which serves as the target against which every uncertainty score
is evaluated. An alternative convention assigns the label by majority vote over the
sampled generations; this is an equally reasonable choice, and
App.~\ref{supp:majority_vote} reports the same experiments under it, with unchanged
conclusions.

\textbf{Metrics.} We report AUROC and ECE. AUROC measures how well an uncertainty score ranks incorrect
generations above correct ones, with $0.5$ corresponding to random ordering and higher
values better; it is invariant to any monotone rescaling of the score, and so evaluates
discrimination alone. ECE measures how far predicted confidences are from empirical
correctness frequencies, averaged over confidence bins, and is lower when the two agree.
Since most of the estimators considered here output an unbounded score rather than a
probability, ECE is computed after mapping scores to confidences with a transformation
fitted on a held-out split, identically for every method.

\subsection{Main results}

Tab.~\ref{tab:main} reports every estimator in its published form and with our operator
substituted, averaged over the 18 model--dataset combinations and broken down by dataset
and by model. 
\begin{table*}[t!]
\caption{\textbf{Main results.} AUROC ($\uparrow$) and ECE ($\downarrow$) for each estimator in its published form
(\emph{base}) and with our operator substituted (\emph{shaded}), averaged over the 18
model--dataset combinations and reported by dataset (top) and by model (bottom).
$\Delta$ is the change in mean AUROC over all columns.}
\label{tab:main}
\centering
\resizebox{\textwidth}{!}{%
\begin{tabular}{c cc cc cc | cc cc cc | c c}
\toprule
& \multicolumn{2}{c}{\textbf{ADVQA~}} & \multicolumn{2}{c}{\textbf{OKVQA~}}
& \multicolumn{2}{c}{\textbf{VizWiz~}} & \multicolumn{2}{c}{\textbf{HotpotQA}}
& \multicolumn{2}{c}{\textbf{TriviaQA}} & \multicolumn{2}{c}{\textbf{WebQuestions}}
&  &  post-hoc\\
Method & AUROC & ECE & AUROC & ECE & AUROC & ECE & AUROC & ECE & AUROC & ECE & AUROC & ECE
& $\Delta$ & compute \\
\midrule
CAE & .637 & .116 & .643 & .229 & .647 & .199 & .673 & .235 & .583 & .212 & .490 & .176 & & 1.11 s \\
\ours $\mathrm{CAE}_{\mathrm{Enc}}$ & \textbf{.663} & \textbf{.081} & \textbf{.733} & \textbf{.194} & \textbf{.674} & \textbf{.159} & \textbf{.722} & \textbf{.215} & \textbf{.840} & \textbf{.135} & \textbf{.716} & \textbf{.055} & +.113 & \textbf{0.03 s} \\
\midrule
SE & .637 & .114 & .643 & .228 & .644 & .198 & .674 & .235 & .577 & .212 & .488 & .175 & & 1.18 s \\
\ours $\mathrm{SE}_{\mathrm{Enc}}$ & \textbf{.666} & \textbf{.080} & \textbf{.733} & \textbf{.193} & \textbf{.675} & \textbf{.159} & \textbf{.725} & \textbf{.214} & \textbf{.840} & \textbf{.133} & \textbf{.719} & \textbf{.058} & +.116 & \textbf{0.03 s} \\
\midrule
KLE-Heat & .675 & .080 & .678 & \textbf{.187} & .690 & \textbf{.161} & .719 & \textbf{.208} & .747 & \textbf{.127} & .555 & .112 & & 1.21 s \\
\ours $\mathrm{KLE\text{-}Heat}_{\mathrm{Enc}}$ & \textbf{.683} & \textbf{.077} & \textbf{.763} & .203 & \textbf{.732} & .169 & \textbf{.763} & .220 & \textbf{.855} & .140 & \textbf{.739} & \textbf{.045} & +.079 & \textbf{0.04 s} \\
\midrule
KLE-Matern & \textbf{.688} & \textbf{.070} & .697 & \textbf{.202} & .712 & .178 & .715 & .226 & .729 & .149 & .548 & .138 & & 1.23 s \\
\ours $\mathrm{KLE\text{-}Matern}_{\mathrm{Enc}}$ & .684 & .074 & \textbf{.763} & .205 & \textbf{.733} & \textbf{.173} & \textbf{.763} & \textbf{.223} & \textbf{.855} & \textbf{.141} & \textbf{.739} & \textbf{.044} & +.075 & \textbf{0.04 s} \\
\midrule
COS-int & .577 & .117 & .621 & \textbf{.203} & .641 & \textbf{.171} & .720 & .228 & .658 & .139 & .602 & .144 & & \textbf{0.01 s} \\
COS-ext & .651 & .094 & .673 & .205 & .705 & .179 & .707 & .230 & .785 & \textbf{.129} & .554 & .135 & & 0.03 s \\
\ours $\mathrm{COS}_{\mathrm{Enc}}$ & \textbf{.694} & \textbf{.072} & \textbf{.758} & .208 & \textbf{.741} & .178 & \textbf{.765} & \textbf{.225} & \textbf{.852} & .141 & \textbf{.741} & \textbf{.043} & +.079 & 0.03 s \\
\midrule
TE & .608 & .099 & .594 & .205 & .645 & .179 & \textbf{.738} & .221 & .669 & .142 & .589 & .121 & & \textbf{0.01 s} \\
\ours $\mathrm{TE}_{\mathrm{Enc}}$ & \textbf{.636} & \textbf{.079} & \textbf{.641} & \textbf{.199} & \textbf{.668} & \textbf{.171} & .699 & \textbf{.216} & \textbf{.769} & \textbf{.119} & \textbf{.655} & \textbf{.074} & +.038 & 0.02 s \\
\bottomrule
\end{tabular}}

\resizebox{\textwidth}{!}{%
\begin{tabular}{c cc cc cc | cc cc cc | c c}
\toprule
& \multicolumn{2}{c}{\texttt{idefics2-8B}} & \multicolumn{2}{c}{\texttt{Qwen2.5-VL-7B}}
& \multicolumn{2}{c}{\texttt{llava-1.5-7B}} & \multicolumn{2}{c}{\texttt{Qwen2.5-7B}}
& \multicolumn{2}{c}{\texttt{Phi-3.5-mini}} & \multicolumn{2}{c}{\texttt{Mistral-7B}}
&  & post-hoc \\
Method & AUROC & ECE & AUROC & ECE & AUROC & ECE & AUROC & ECE & AUROC & ECE & AUROC & ECE
& $\Delta$ & compute \\
\midrule
CAE & .654 & .192 & .626 & .196 & .647 & .156 & .639 & .179 & .517 & .232 & .590 & .213 & & 1.11 s \\
\ours $\mathrm{CAE}_{\mathrm{Enc}}$ & \textbf{.678} & \textbf{.159} & \textbf{.717} & \textbf{.144} & \textbf{.675} & \textbf{.131} & \textbf{.769} & \textbf{.132} & \textbf{.764} & \textbf{.130} & \textbf{.744} & \textbf{.142} & +.112 & \textbf{0.03 s} \\
\midrule
SE & .654 & .189 & .624 & .195 & .645 & .155 & .638 & .178 & .515 & .231 & .586 & .213 & & 1.18 s \\
\ours $\mathrm{SE}_{\mathrm{Enc}}$ & \textbf{.679} & \textbf{.158} & \textbf{.717} & \textbf{.144} & \textbf{.678} & \textbf{.130} & \textbf{.770} & \textbf{.131} & \textbf{.767} & \textbf{.130} & \textbf{.747} & \textbf{.143} & +.116 & \textbf{0.03 s} \\
\midrule
KLE-Heat & .687 & .162 & .672 & \textbf{.145} & .684 & \textbf{.121} & .698 & .138 & .629 & .154 & .693 & .155 & & 1.21 s \\
\ours $\mathrm{KLE\text{-}Heat}_{\mathrm{Enc}}$ & \textbf{.723} & \textbf{.161} & \textbf{.740} & .155 & \textbf{.715} & .134 & \textbf{.792} & \textbf{.127} & \textbf{.792} & \textbf{.132} & \textbf{.774} & \textbf{.145} & +.079 & \textbf{0.04 s} \\
\midrule
KLE-Matern & .707 & .165 & .681 & .156 & .709 & \textbf{.130} & .693 & .148 & .611 & .183 & .689 & .182 & & 1.23 s \\
\ours $\mathrm{KLE\text{-}Matern}_{\mathrm{Enc}}$ & \textbf{.724} & \textbf{.163} & \textbf{.740} & \textbf{.155} & \textbf{.716} & .133 & \textbf{.791} & \textbf{.128} & \textbf{.792} & \textbf{.134} & \textbf{.774} & \textbf{.147} & +.075 & \textbf{0.04 s} \\
\midrule
COS-int & .592 & .168 & .625 & .185 & .623 & .139 & .699 & .168 & .605 & .204 & .685 & \textbf{.138} & & \textbf{0.01 s} \\
COS-ext & .669 & .174 & .693 & .162 & .667 & .142 & .688 & .157 & .704 & .137 & .653 & .199 & & 0.03 s \\
\ours $\mathrm{COS}_{\mathrm{Enc}}$ & \textbf{.730} & \textbf{.165} & \textbf{.740} & \textbf{.159} & \textbf{.723} & \textbf{.133} & \textbf{.790} & \textbf{.124} & \textbf{.793} & \textbf{.135} & \textbf{.775} & .150 & +.080 & 0.03 s \\
\midrule
TE & .615 & .175 & .613 & .172 & .619 & .136 & .716 & .142 & .659 & .146 & .621 & .195 & & \textbf{0.01 s} \\
\ours $\mathrm{TE}_{\mathrm{Enc}}$ & \textbf{.650} & \textbf{.164} & \textbf{.644} & \textbf{.154} & \textbf{.651} & \textbf{.131} & \textbf{.741} & \textbf{.123} & \textbf{.714} & \textbf{.125} & \textbf{.668} & \textbf{.161} & +.038 & 0.02 s \\
\bottomrule
\end{tabular}}
\scriptsize \textit{Note.} Compute is the post-hoc time to build $\mS$ for one question; excluding the generation shared by all methods we compare against.

\end{table*}

\textbf{Replacing the comparison operator improves every semantic estimator.} Substituting the learned operator raises the mean AUROC for every baseline estimator across the 18 model--dataset combinations: CAE improves from $0.612$ to $0.724$, SE from $0.610$ to $0.726$, KLE-Heat from $0.677$ to $0.756$, KLE-Matern from $0.682$ to $0.756$, and COS from $0.679$ (for the stronger variant) to $0.759$. Overall 120 of 126 comparisons (95\%), obtained by pairing each of the 7 estimators with its operator-substituted counterpart across all 18 model--dataset combinations spanning LLMs and LVLMs, show an improvement. Per-cell results with bootstrapped CIs are detailed in Tabs.~\ref{tab:percell_text} and \ref{tab:percell_vqa}.

\textbf{The operator is substantially cheaper.} Building $\mS$ with the conventional NLI operator costs $1.1$--$1.2$,s per question; with ours, it takes $0.03$--$0.04$,s, representing roughly a $30\times$ reduction, since the comparison no longer requires a forward pass per pair. COS-int is the only operator cheaper still, at the cost of white-box access and yielding the weakest performance in the table.

\textbf{The question matters.} Tab.~\ref{tab:main} reports $\mathrm{COS}_{\mathrm{Enc}}$ with the question as part of the encoder input, as described in Sec.~\ref{sec:method}. Retraining the same operator on answer strings alone, removing \quentin{$\vx_{\mathrm{text}}$ from the input of $E_{\vpsi}$}, drops the mean AUROC from $0.759$ to $0.742$. The gap is smaller than the gain over the conventional operators, but it holds in every one of the 18 model--dataset combinations (Tab.~\ref{question_ablation}): conditioning is a real, if secondary, part of the improvement, and consistent with \textbf{(D2)}, which no operator defined on answer pairs alone can satisfy.

\textbf{The token-level estimator also improves.} The pattern holds even for the cheapest estimator: reweighting TE with our operator raises the mean AUROC from $0.640$ to $0.678$, using a single generation and one extra encoder pass. This shows the operator is useful beyond semantic UQ.

Additional experiments on the necessity of hard negatives, sensitivity to the clustering threshold, other hyperparameters, and further ablations are provided in App.~\ref{sup:additional_exp}.

\section{Conclusion}

We formalized semantic uncertainty quantification as the composition of an operator that decides whether two generations answer a question the same way, and an aggregator that turns the resulting comparisons into a scalar. We observed that every method in this family proposes the latter while inheriting the former from a model built for another task. We introduced a framework for training such an operator on the relation it is meant to express, utilizing a contrastive objective and synthetic data disjoint from every evaluation setting. Substituted into SE, CAE, KLE, and COS, it improves them across 18 model--dataset combinations, requiring only $N$ network evaluations instead of $O(N^2)$. The uniformity of this improvement substantiates our core hypothesis: a single change to the shared component improves every baseline estimator built on it.

The operator is trained on a small synthetic set with a single configuration, and might benefit from more data and a finer-grained objective. We left it there deliberately: the claim we set out to test is that the operator is the limiting component, not that ours is the best one, and our aim was not to introduce a new supervised uncertainty method. That a recipe this minimal already improves every estimator is the point, and it leaves considerable room above the estimates reported here.

\newpage
\section*{AI use statement}

We used generative AI tools solely to assist with the writing of this manuscript: refining
sentence-level phrasing and style, and formatting tables. Separately, an LLM is used within
the method itself, as a generator and a judge for constructing the training data described
in Sec.~\ref{sec:method}; this is a documented and integral part of the proposed pipeline
rather than an authoring aid, and is evaluated as such throughout the paper. All research
ideas, the formalism, experimental design, code, results, and their interpretation are the
authors' own and were not suggested, generated, or verified by AI tools. We have reviewed
and take full responsibility for every sentence and table in this paper, including any
produced with AI assistance.

\subsection*{Reproducibility statement}

All details needed to reproduce our results are given in the paper. The construction of the
training set, including the source dataset and split, the generating model, sampling
parameters and triplet mining, is described in App.~\ref{supp:training_details}, together
with the training hyperparameters of the operator. The correctness-labelling protocol and
the judges used for training and evaluation are specified in App.~\ref{supp:llm_judge}.
The evaluated models, dataset splits, prompts and decoding parameters are listed in
App.~\ref{supp:eval_details}, which also gives the operators and hyperparameters used for
every baseline. The construction of the diagnostic benchmark of Sec.~\ref{sec:diagnostics}
is detailed in App.~\ref{supp:limit_op}, and per-cell results with bootstrapped confidence
intervals are reported in App.~\ref{supp:additional_results}. All models, encoders and
datasets used are publicly available. Code, trained operator weights and the generated
training data will be released upon acceptance.

\bibliography{iclr2027_conference}
\bibliographystyle{iclr2027_conference}

\appendix

\newpage
\appendix
\numberwithin{equation}{section}
\numberwithin{table}{section}
\numberwithin{figure}{section}

\section{Operators and aggregators}
\label{supp:formalism}

\subsection{{Preliminaries}}

Given an input $\vx$, {consisting of a textual query $\vx_{\mathrm{text}}$ possibly
accompanied by an image $\vx_{\mathrm{im}}$, a model parameterized by $\vtheta$ autoregressively generates an output response sequence of tokens $\vy=\bigl\{\vy_t\bigr\}_{t=1}^T$, where $\vy_t$ denotes the $t^{\text{-th}}$ output token
and $T$ is the length of the output sequence. The probability of the generated sequence $\vy$ is the joint probability of tokens, defined as the product of conditional token probabilities:
\begin{equation}
    p(\vy \mid \vx, \vtheta)=\prod_{t=1}^T p(\vy_t \mid \vx, \vy_{<t}, \vtheta),
    \quad \text{where } \vy_{<t}=\bigl\{\vy_1, \ldots, \vy_{t-1}\bigr\}.
\end{equation}

\textbf{Token Entropy.}
TE is the simplest uncertainty estimate, it operates directly on the predictive distributions produced during decoding, and requires a single generation. At step $t$, the model defines a distribution over the vocabulary $\mathcal{V}$, whose Shannon entropy
\begin{equation}
\label{eq:token_level_entropy}
  H_t(\vx, \vy_{<t}, \vtheta) = -\sum_{v \in \mathcal{V}}
  p(v \mid \vx, \vy_{<t}, \vtheta) \log p(v \mid \vx, \vy_{<t}, \vtheta)
\end{equation}
measures how dispersed the model is over the possible continuations at that position. The token-level uncertainty of a generation is obtained by averaging these entropies over the generated sequence:
\begin{equation}
  U_{\mathrm{TE}}(\vy \mid \vx, \vtheta) = \frac{1}{T} \sum_{t=1}^{T}
  H_t(\vx, \vy_{<t}, \vtheta).
  \label{eq:token_entropy}
\end{equation}
This estimate is cheap, requiring neither sampling nor comparison between generations, but it is computed entirely in token space. Every position contributes equally to the average, although the positions differ widely in how much they bear on the answer: a model may hesitate between several equally acceptable phrasings while being certain of the answer itself, and such hesitation raises $U_{\mathrm{TE}}$ exactly as hesitation over the answer would.
Semantic methods compare generations by meaning rather than by token sequence.

This section reviews these methods and isolates the component they share. We first introduce the comparison matrix every such method is built on and the forms it takes in the literature (Sec.~\ref{sec:comparison_matrix}), then present the estimators themselves (Sec.~\ref{sec:estimators}), and formalize their decomposition (Sec.~\ref{sec:operator_aggregator_details}).

\subsection{The comparison matrix}
\label{sec:comparison_matrix}

Each semantic estimator starts from a pairwise comparison of the sampled generations
$\mathcal{Y}:=\{\vy^{(1)}, \ldots, \vy^{(N)}\}$, collected in a matrix
\begin{equation}
  \mS \in \R^{N \times N}, \qquad
  \mS_{ij} = s\bigl(\vy^{(i)}, \vy^{(j)} \mid \vx_{\mathrm{text}}\bigr),
  \label{eq:S}
\end{equation}
where the \emph{operator} $s$ scores how far two generations express the same answer to
$\vx_{\mathrm{text}}$. Two instantiations of $s$ account for every method considered here.

\textbf{Cosine similarity ($\mS^{\mathrm{cos}}$).}
Each generation is mapped to a normalized embedding $\vphi_i \in \R^d$ and compared by
inner product, $\mS^{\mathrm{cos}}_{ij} = \langle \vphi_i, \vphi_j \rangle$. The
embeddings are taken either from the hidden states of the generating model
(COS-int)~\citet{chen2024, binkowski2025} or from an external sentence
encoder (COS-ext)~\citet{grewal2024improving, abdaljalil2025}.

\textbf{Bidirectional entailment ($\mS^{\mathrm{nli}}$).}
Each generation is concatenated with $\vx_{\mathrm{text}}$ and the pair is passed to a pretrained NLI
model, which predicts one of entailment, neutral or contradiction. The operator records
whether entailment is predicted in both directions,

\begin{equation}
  \mS^{\mathrm{nli}}_{ij} = \mathbf{1}\bigl[\vx_{\mathrm{text}} \land \vy^{(i)} \models \vy^{(j)}
  \ \text{ and } \ \vx_{\mathrm{text}} \land \vy^{(j)} \models \vy^{(i)}\bigr] \in \{0,1\}.
  \label{eq:S_nli}
\end{equation}
Some methods instead retain the entailment probabilities underlying these predictions,
giving a graded matrix $\mS^{\mathrm{nli}}_{\mathrm{soft}} \in [0,1]^{N \times N}$ that
distinguishes pairs the hard decision treats alike.

In both cases the matrix is inherited from a component developed for another purpose.
The estimators below draw on one or the other and differ in what they compute from it.

\subsection{Semantic estimators}
\label{sec:estimators}

\textbf{Semantic clustering.}
SE and CAE both reduce the comparison to a partition of $\mathcal{Y}$ into semantically
consistent clusters $\mathcal{C}:=\{\mathcal{C}_k\}_{k=1}^K$, such that sequences within a
cluster share the same meaning. They consume $\mS^{\mathrm{nli}}$:
clusters are formed by comparing each new generation against the existing ones and
creating a new cluster when no bidirectional entailment is
found~\citet{kuhnsemantic, farquhar2024detecting}.

\textbf{Semantic Entropy.}
SE weights each cluster by the normalized sequence probabilities of its members,
\begin{equation}
    p(\mathcal{C}_k \mid \vx, \vtheta) = \sum_{\vy \in \mathcal{C}_k}
    \tilde p(\vy \mid \vx, \vtheta),
    \qquad
    \tilde p(\vy \mid \vx, \vtheta)=\frac{p(\vy \mid \vx, \vtheta)}
    {\sum_{i=1}^N p(\vy^{(i)} \mid \vx, \vtheta)},
    \label{eq:proba_cluster}
\end{equation}
and returns the entropy of the resulting cluster distribution:
\begin{equation}
    H_{\mathrm{SE}}(\mathcal{Y} \mid \vx, \vtheta) = - \sum_{k=1}^K
    p(\mathcal{C}_k \mid \vx, \vtheta) \log p(\mathcal{C}_k \mid \vx, \vtheta).
    \label{eq:semantic_entropy}
\end{equation}
A low value indicates that the responses concentrate on a single meaning, a high value that probability mass is dispersed across distinct meanings.

\textbf{Cluster-Assignment Entropy.}
CAE uses the same partition but replaces the probability mass of each cluster by its empirical frequency, removing the need for access to sequence probabilities~\citet{farquhar2024detecting}:
\begin{equation}
  H_{\mathrm{CAE}}(\mathcal{Y} \mid \vx)
  = -\sum_{k=1}^{K} \frac{|\mathcal{C}_k|}{N} \log \frac{|\mathcal{C}_k|}{N}.
  \label{eq:cae}
\end{equation}

\textbf{Kernel Language Entropy.}
Rather than reducing the comparisons to a partition, KLE turns them into a weighted
semantic graph over the generations, taking the graded matrix as edge weights,
$\mW = \mS^{\mathrm{nli}}_{\mathrm{soft}}$~\citet{nikitin2024kernel}. From its Laplacian
$\mL = \mD - \mW$, where $\mD$ is the diagonal degree matrix
$\mD_{ii} = \sum_{j} \mW_{ij}$, a graph kernel is obtained as the solution of a diffusion
equation,
$\mK_t = e^{-t\mL}$ for the heat kernel (KLE-Heat) or
$\mK_{\nu\kappa} = (2\nu/\kappa^2 \mI + \mL)^{-\nu}$ for the Mat\'ern kernel
(KLE-Matern); after normalization to unit trace, the uncertainty is its von Neumann
entropy,
\begin{equation}
  H_{\mathrm{KLE}}(\mathcal{Y} \mid \vx) = -\operatorname{Tr}\bigl(\mK \log \mK\bigr).
  \label{eq:kle}
\end{equation}
Because the graph carries graded rather than binary weights, KLE distinguishes
generations that are related without being equivalent, which hard clustering cannot; it
is in fact a strict generalization of SE, which it recovers for a block-diagonal
kernel.

\textbf{Eigenvalue-based scores.}
COS-int and COS-ext consume $\mS^{\mathrm{cos}}$ and summarize its spectrum directly,
differing only in the embeddings it is built from. Being a Gram
matrix of normalized vectors, $\mS^{\mathrm{cos}}$ is positive semidefinite with trace
$N$, so its eigenvalues
$\lambda_1 \geq \cdots \geq \lambda_N$ normalized by their sum form a distribution over
the orthogonal directions of the embedding space. The uncertainty is the von Neumann
entropy of that distribution~\citet{chen2024}:
\begin{equation}
  U_{\mathrm{COS}}(\mathcal{Y} \mid \vx) = -\sum_{i=1}^{N}
  \tilde\lambda_i \log \tilde\lambda_i,
  \qquad \tilde\lambda_i = \frac{\lambda_i}{\sum_{j=1}^{N} \lambda_j} =
  \frac{\lambda_i}{N}.
  \label{eq:cos}
\end{equation}
A spectrum dominated by one eigenvalue indicates that the generations concentrate on a single direction of meaning, a flat spectrum that they are dispersed across several.

\subsection{{Decomposing Existing Semantic UQ Methods into an Operator and an Aggregator}}
\label{sec:operator_aggregator_details}

\paragraph{A common view.}
Although semantic UQ methods look different, most of them follow the same
two-step structure.
Given a set of sampled generations
\(
    \mathcal{Y}
    =
    \{\vy^{(1)},\ldots,\vy^{(N)}\},
\)
they first compare the generations pairwise through an \emph{operator} $s$,
producing
\begin{equation}
    \mS_{ij}
    =
    s
    \left(
        \vy^{(i)},\vy^{(j)}
        \mid
        \vx_{\mathrm{text}}
    \right).
    \label{eq:common_operator}
\end{equation}
They then apply an \emph{aggregator} $F$ to these pairwise relations,
\begin{equation}
    \boxed{
    U = F(\mS).
    }
    \label{eq:common_operator_aggregator}
\end{equation}
For methods that also use sequence probabilities, we write more generally
\begin{equation}
    U
    =
    F
    \left(
        \mS,
        \left\{
            p(\vy^{(i)}\mid\vx,\vtheta)
        \right\}_{i=1}^{N}
    \right).
\end{equation}

The distinction is simple:
the operator $s$ decides \emph{how two answers are related}, while the
aggregator $F$ decides \emph{how all these relations are combined into one
uncertainty score}.
Below, we make both $s$ and $F$ explicit for each estimator.

\subsubsection{Semantic Entropy}

\paragraph{SE: NLI operator followed by probability-weighted cluster entropy.}
Semantic Entropy first compares the generations using a binary NLI relation.
Its operator is
\begin{equation}
    s_{\mathrm{SE}}
    =
    s_{\mathrm{NLI}},
\end{equation}
which produces the binary matrix
\(
    \mS^{\mathrm{nli}}.
\)

Let
\begin{equation}
    \mathcal{C}(\mS)
    =
    \Pi(\mS)
    =
    \{
        \mathcal{C}_1(\mS),\ldots,\mathcal{C}_{K(\mS)}(\mS)
    \}
\end{equation}
denote the semantic partition obtained by applying the clustering rule
$\Pi$ to the matrix $\mS$.
For each cluster, define
\begin{equation}
    q_k(\mS)
    =
    \sum_{\vy^{(i)}\in\mathcal{C}_k(\mS)}
    \tilde p_i,
    \qquad
    \tilde p_i
    =
    \frac{
        p(\vy^{(i)}\mid\vx,\vtheta)
    }{
        \sum_{j=1}^{N}
        p(\vy^{(j)}\mid\vx,\vtheta)
    }.
\end{equation}

The SE aggregator is therefore the function
\begin{equation}
    \boxed{
    F_{\mathrm{SE}}
    \left(
        \mS,\{\tilde p_i\}_{i=1}^{N}
    \right)
    :=
    -\sum_{k=1}^{K(\mS)}
    q_k(\mS)\log q_k(\mS).
    }
    \label{eq:F_se}
\end{equation}
Thus,
\begin{equation}
    U_{\mathrm{SE}}
    =
    F_{\mathrm{SE}}
    \left(
        \mS^{\mathrm{nli}},
        \{\tilde p_i\}_{i=1}^{N}
    \right).
\end{equation}

The operator determines which responses belong to the same semantic
cluster, while $F_{\mathrm{SE}}$ measures how the sequence probability mass
is distributed across these clusters.

\subsubsection{Cluster-Assignment Entropy}

\paragraph{CAE: the same semantic operator with a different aggregator.}
CAE uses the same binary NLI matrix and the same clustering rule,
\begin{equation}
    \mathcal{C}(\mS)
    =
    \Pi(\mS).
\end{equation}
The difference from SE comes after the clusters have been obtained.
CAE assigns to each cluster its empirical frequency,
\begin{equation}
    \widehat q_k(\mS)
    =
    \frac{
        |\mathcal{C}_k(\mS)|
    }{N}.
\end{equation}

Its aggregator is therefore
\begin{equation}
    \boxed{
    F_{\mathrm{CAE}}(\mS)
    :=
    -\sum_{k=1}^{K(\mS)}
    \widehat q_k(\mS)
    \log
    \widehat q_k(\mS).
    }
    \label{eq:F_cae}
\end{equation}
Hence,
\begin{equation}
    U_{\mathrm{CAE}}
    =
    F_{\mathrm{CAE}}
    \left(
        \mS^{\mathrm{nli}}
    \right).
\end{equation}

SE and CAE therefore use the same semantic operator and clustering rule.
They differ only in the final aggregation:
SE uses sequence probability mass, while CAE uses empirical cluster
frequency.

\subsubsection{Kernel Language Entropy}

\paragraph{KLE: continuous NLI operator followed by graph aggregation.}
KLE does not first reduce semantic relations to binary clusters.
Instead, its operator produces a continuous semantic matrix
\begin{equation}
    \mS
    =
    \mS^{\mathrm{nli}}_{\mathrm{soft}},
\end{equation}
which is interpreted as the weighted adjacency matrix
\begin{equation}
    \mW(\mS)=\mS.
\end{equation}
From this matrix, define
\begin{equation}
    \mD(\mS)_{ii}
    =
    \sum_j \mW(\mS)_{ij},
    \qquad
    \mL(\mS)
    =
    \mD(\mS)-\mW(\mS).
\end{equation}

For KLE-Heat, the kernel is
\begin{equation}
    \mK_{\mathrm{Heat}}(\mS)
    =
    \exp
    \left(
        -t\mL(\mS)
    \right),
\end{equation}
while for KLE-Matern,
\begin{equation}
    \mK_{\mathrm{Matern}}(\mS)
    =
    \left(
        \frac{2\nu}{\kappa^2}\mI
        +
        \mL(\mS)
    \right)^{-\nu}.
\end{equation}

For either kernel, define its trace-normalized form
\begin{equation}
    \bar{\mK}(\mS)
    =
    \frac{
        \mK(\mS)
    }{
        \operatorname{Tr}(\mK(\mS))
    }.
\end{equation}
The KLE aggregator is then
\begin{equation}
    \boxed{
    F_{\mathrm{KLE}}(\mS;\mK)
    :=
    -
    \operatorname{Tr}
    \left(
        \bar{\mK}(\mS)
        \log
        \bar{\mK}(\mS)
    \right).
    }
    \label{eq:F_kle}
\end{equation}

More explicitly,
\begin{align}
    F_{\mathrm{Heat}}(\mS)
    &:=
    -
    \operatorname{Tr}
    \left(
        \bar{\mK}_{\mathrm{Heat}}(\mS)
        \log
        \bar{\mK}_{\mathrm{Heat}}(\mS)
    \right),
    \\
    F_{\mathrm{Matern}}(\mS)
    &:=
    -
    \operatorname{Tr}
    \left(
        \bar{\mK}_{\mathrm{Matern}}(\mS)
        \log
        \bar{\mK}_{\mathrm{Matern}}(\mS)
    \right).
\end{align}
Therefore,
\begin{align}
    U_{\mathrm{KLE\text{-}Heat}}
    &=
    F_{\mathrm{Heat}}
    \left(
        \mS^{\mathrm{nli}}_{\mathrm{soft}}
    \right),
    \\
    U_{\mathrm{KLE\text{-}Matern}}
    &=
    F_{\mathrm{Matern}}
    \left(
        \mS^{\mathrm{nli}}_{\mathrm{soft}}
    \right).
\end{align}

The two KLE variants therefore use the same semantic operator.
They differ only in the graph kernel used inside the aggregator.

\subsubsection{COS}

\paragraph{COS: cosine operator followed by spectral aggregation.}
COS builds its comparison matrix from cosine similarities between answer
representations,
\begin{equation}
    \mS^{\mathrm{cos}}_{ij}
    =
    \left\langle
        \vphi_i,\vphi_j
    \right\rangle.
\end{equation}
COS-int and COS-ext differ only in the source of $\vphi_i$:
COS-int uses representations from the generating model, whereas COS-ext
uses an external sentence encoder.

For any positive semidefinite similarity matrix $\mS$, let
\begin{equation}
    \lambda_1(\mS),\ldots,\lambda_N(\mS)
\end{equation}
be its eigenvalues and define
\begin{equation}
    \widetilde\lambda_i(\mS)
    =
    \frac{
        \lambda_i(\mS)
    }{
        \sum_j\lambda_j(\mS)
    }
    =
    \frac{
        \lambda_i(\mS)
    }{
        \operatorname{Tr}(\mS)
    }.
\end{equation}

The COS aggregator is
\begin{equation}
    \boxed{
    F_{\mathrm{COS}}(\mS)
    :=
    -
    \sum_{i=1}^{N}
    \widetilde\lambda_i(\mS)
    \log
    \widetilde\lambda_i(\mS).
    }
    \label{eq:F_cos}
\end{equation}
Hence,
\begin{equation}
    U_{\mathrm{COS}}
    =
    F_{\mathrm{COS}}
    \left(
        \mS^{\mathrm{cos}}
    \right).
\end{equation}

Since the embeddings are normalized,
$\operatorname{Tr}(\mS^{\mathrm{cos}})=N$.
The operator therefore defines the geometry of the generations, while
$F_{\mathrm{COS}}$ measures how this geometry is spread across its spectral
directions.

\section{Additional Experiments}
\label{sup:additional_exp}
\subsection{Clustering of Incorrect Answers}
\label{supp:wrong_answer_clustering}

As described in Section~\ref{sec:contrastive_objective}, we use answer correctness as a proxy for semantic equivalence when building training triplets. Two answers to the same question form a positive pair if both are correct, and a negative pair if one is correct and the other is not. Pairs of two incorrect answers therefore never appear as positives during training. This raises a natural concern: the encoder might only learn to group correct answers together, and fail to group incorrect answers that share the same meaning.

We argue that this concern does not hold, because correctness enters the pipeline only through the construction of triplets. The encoder never receives a correctness label as input and is never trained to predict one. The objective in Eq.~\ref{eq:triplet_loss} depends only on inner products between answer embeddings, so what the encoder learns is a notion of \emph{equivalence} between answers, not of \emph{truth}. For a given question, all correct answers share the same denotation, which makes correctness a convenient and reliable source of equivalent pairs. The notion of equivalence the encoder learns from these pairs is not tied to the correct answer, and it transfers to answers that agree on the same incorrect content.

Tab.~\ref{tab:wrong_answer_example} illustrates this behavior with two pairs of paraphrases answering the same question: one pair is correct and the other is incorrect. The encoder assigns nearly identical similarities to both pairs ($0.975$ and $0.980$). It therefore groups paraphrases based on their shared meaning, regardless of whether that meaning is correct. This is precisely the behavior required by downstream semantic UQ aggregators: when a model consistently produces the same hallucination, the corresponding samples should fall into a single cluster, yielding low semantic entropy that reflects the model's (misplaced) confidence.

\begin{table}[h!]
  \centering
  \caption{Similarity assigned by our encoder to pairs of semantically equivalent answers, for correct and incorrect content.
  Question: \emph{``What is the capital of France?''}}
  \label{tab:wrong_answer_example}
  \begin{tabular}{lllc}
    \toprule
    Correctness & Answer 1 & Answer 2 & Similarity \\
    \midrule
    Both correct   & The capital of France is Paris.  & Paris is the capital of France.  & .975 \\
    Both incorrect & The capital of France is Madrid. & Madrid is the capital of France. & .980 \\
    \bottomrule
  \end{tabular}
\end{table}

\subsection{Sensitivity to base encoder}
\label{sec:backbone}

{
A natural question is whether the gains come from the particular encoder we
fine-tune or from the training framework itself.
We therefore repeat the procedure of Sec.~\ref{sec:method} with six public
sentence encoders, covering different model sizes and pretraining strategies.
For every backbone, we keep the training data, objective, margin, and number
of epochs unchanged, and evaluate the resulting
$\mathrm{COS}_{\mathrm{Enc}}$.
}

{Tab.~\ref{tab:backbone} reports the mean AUROC over the 18 model–dataset combinations.
The difference between the six trained backbones is small, only $0.009$,
and even the weakest trained encoder outperforms the best encoder before
fine-tuning.
Moreover, larger models are not consistently better, and the ranking of the
encoders changes after training.
These results suggest that the main gain does not come from a particular
backbone or from simply improving an already strong representation.
Instead, our training adds a question-conditioned notion of answer
equivalence that transfers across different encoder architectures.
}

\begin{table}[h!]
\centering
\caption{\textbf{The framework is insensitive to the backbone.} Mean AUROC of
$\mathrm{COS}_{\mathrm{Enc}}$ over the 18 model--dataset combinations, for six sentence
encoders used as the starting point of the training procedure of Sec.~\ref{sec:method},
with all other settings held fixed.}
\label{tab:backbone}
\resizebox{0.5\textwidth}{!}{
\begin{tabular}{lc}
\toprule
Backbone & AUROC \\
\midrule
\ours \texttt{sentence-t5-large}          & .759 \\
\texttt{all-roberta-large-v1}       & .759 \\
\texttt{all-mpnet-base-v2}          & .756 \\
\texttt{multi-qa-mpnet-base-dot-v1} & .755 \\
\texttt{bge-base-en-v1.5}           & .752 \\
\texttt{gtr-t5-large}               & .750 \\
\bottomrule
\end{tabular}
}
\end{table}

\subsection{Negatives are necessary, not incidental}
\label{sec:pair_ablation}

{The objective in Eq.~(\ref{eq:triplet_loss}) has two goals: bring equivalent
answers closer and push different answers apart.
This raises a simple question: \textit{are negative examples really needed, and does
their choice matter?}}

{In our setting, the negatives are particularly informative.
They are not unrelated sentences, but incorrect answers to the \emph{same}
question, generated by the same model.
The anchor and the negative therefore share the same topic and often a similar
structure, while differing mainly in the answer itself.
To separate them, the encoder cannot rely only on topic, style, or length.
It must focus on the part of the response that carries the answer.
We therefore refer to these examples as \emph{hard negatives}.
}

{To study their role, we compare four settings using the same encoder and
training questions.
The first is the \textit{untrained encoder}.
The second uses\textit{ only positive pairs, with no negative term}.
The third uses\textit{ random negatives, sampled from answers to different
questions}.
The last uses \textit{the hard negatives of our method, sampled from incorrect
answers to the same question as the anchor}.
}

{
Tab.~\ref{tab:pair} shows a clear difference.
The mean AUROC is $0.679$ for the untrained encoder, $0.650$ without negatives,
$0.705$ with random negatives, and $0.759$ with hard negatives.
Training only on positive pairs even performs worse than the untrained
encoder, suggesting that pulling examples together without any repulsive term
can lead to a less informative representation.
Random negatives improve the result, showing that pushing different examples
apart is important.
However, they remain clearly below hard negatives: answers from different
questions can often be separated using topic alone.
In contrast, same-question negatives share the topic and differ mainly in the
answer, forcing the encoder to learn factual differences.
This explains why our training uses negatives from the same question rather
than random responses from the full training set.
}

\begin{table}[t]
\centering
\small
\setlength{\tabcolsep}{5pt}
\caption{\textbf{Hard negatives matter, not just negatives.} AUROC of
$\mathrm{COS}_{\mathrm{Enc}}$ for the untrained encoder and under three training regimes:
no negatives (cosine-similarity loss on positive pairs only), random negatives (drawn from
a different question), and hard negatives (drawn from the same question, used throughout
this paper), averaged over models (top) and over datasets (bottom).}
\label{tab:pair}
\resizebox{1\textwidth}{!}{
\begin{tabular}{lcccccc|c}
\toprule
& \textbf{ADVQA} & \textbf{OKVQA} & \textbf{VizWiz} & \textbf{HotpotQA}
& \textbf{TriviaQA} & \textbf{WebQuestions} & \textbf{Mean} \\
\midrule
\ours hard negatives (ours)
& \textbf{.694} & \textbf{.758} & \textbf{.741} & \textbf{.765} & \textbf{.852}
& \textbf{.741} & \textbf{.759} \\
random negatives
& .666 & .690 & .697 & .738 & .823 & .618 & .705 \\
untrained encoder
& .651 & .673 & .705 & .707 & .785 & .554 & .679 \\
no negatives
& .585 & .616 & .610 & .687 & .760 & .640 & .650 \\
\midrule
& \texttt{idefics2-8B} & \texttt{Qwen2.5-VL-7B} & \texttt{llava-1.5-7B}
& \texttt{Qwen2.5-7B} & \texttt{Phi-3.5-mini} & \texttt{Mistral-7B} & \textbf{Mean} \\
\midrule
\ours hard negatives (ours)
& \textbf{.730} & \textbf{.740} & \textbf{.723} & \textbf{.790} & \textbf{.793}
& \textbf{.775} & \textbf{.759} \\
random negatives
& .685 & .690 & .679 & .738 & .734 & .707 & .705 \\
untrained encoder
& .669 & .693 & .667 & .688 & .704 & .653 & .679 \\
no negatives
& .596 & .628 & .588 & .711 & .740 & .636 & .650 \\
\bottomrule
\end{tabular}
}
\end{table}

\subsection{Sensitivity to Training Hyperparameters}
\label{app:hp_sensitivity}

{The operator is trained with a margin of $0.2$ and a learning rate of
$2\times10^{-5}$, fixed once and used in all main experiments.
To test whether our results depend strongly on these choices, we vary the
margin and learning rate separately while keeping all other settings unchanged.
}

{Tab.~\ref{tab:hp} reports the AUROC of $\mathrm{COS}_{\mathrm{Enc}}$ for the
different configurations.
The results vary only slightly, from $0.749$ to $0.765$.
Even the weakest setting still improves over the untrained encoder by $0.070$.
The hyperparameters used in the main paper are also not the best-performing
ones: using a margin of $0.1$ improves AUROC by $0.006$.
We nevertheless keep the original configuration, since tuning the
hyperparameters on the evaluation settings would make the comparison less
fair.
}

\begin{table}[h!]
\centering
\caption{\textbf{Small variations across hyperparameters.} AUROC of
$\mathrm{COS}_{\mathrm{Enc}}$, averaged over the 18 model--dataset combinations, when the
margin and learning rate are varied around the configuration used throughout
(shaded). All other settings are held fixed.}
\label{tab:hp}
\begin{tabular}{ccc}
\toprule
Margin & Learning rate & AUROC \\
\midrule
0.10 & $2\times10^{-5}$ & .765 \\
0.15 & $2\times10^{-5}$ & .758 \\
\ours 0.20 & $2\times10^{-5}$ & .759 \\
0.30 & $2\times10^{-5}$ & .757 \\
0.50 & $2\times10^{-5}$ & .749 \\
\midrule
0.20 & $1\times10^{-5}$ & .759 \\
0.20 & $3\times10^{-5}$ & .761 \\
0.20 & $5\times10^{-5}$ & .758 \\
\bottomrule
\end{tabular}
\end{table}

\subsection{Token-Level Localization of the Ground Truth}
\label{app:token_localization}

{In Sec.~\ref{sec:token_level}, we interpret the token norm
$\|\vh_t\|_2$ as a measure of how much token $t$ contributes to the semantic
representation of the answer.
This interpretation leads to a simple prediction:
if large norms really identify the most important parts of the answer, then
the highest-weighted token should often lie on the ground-truth answer itself.
We test this directly.}

{We use \texttt{Mistral-7B} on the \textbf{TriviaQA} val set.
We keep only greedy generations that are labelled correct and contain exactly
one annotated ground-truth alias as a substring.
This leaves 232 examples out of 1,000.
For each example, we compute the token weights $w_t$ on the generated answer,
conditioned on the question, and match the ground-truth text span to the
encoder tokens using character offsets.}

{The token with the largest weight falls inside the ground-truth span in
167 out of 232 cases, corresponding to $72\%$.
For comparison, a token drawn uniformly at random from the answer would fall
inside the ground-truth span only $3.9\%$ of the time.
This corresponds to an improvement of roughly $18\times$ over chance.
Importantly, the encoder receives no token-level supervision during training:
the triplet objective of Eq.~\ref{eq:triplet_loss} only compares complete
answers.
The result therefore suggests that the learned representation naturally
localizes the part of the response that carries the factual answer.
}

{Figure~\ref{fig:token_weights} illustrates this behavior on three generations
for the same question.
In each case, most of the weight is placed on the answer token, even when its
position changes: it may appear at the end of the sentence, at the beginning,
or after an unrelated introductory clause.
In contrast, tokens describing the topic of the question receive little
weight.
For example, words such as \texttt{France} and \texttt{capital} are relevant
to the topic, but they are shared by both correct and incorrect answers and
therefore do not help distinguish them.}

{This behavior is consistent with our training objective.
A hard negative answers the same question as the anchor and often shares most
of its wording, while differing mainly in the factual answer.
To separate the two, the encoder must therefore focus on the part of the
sentence that changes the answer, rather than on words that only describe the
general topic.}

\begin{figure}[h!]
\centering
\setlength{\fboxrule}{0.4pt}
\setlength{\fboxsep}{6pt}
\fbox{\includegraphics[width=0.97\textwidth]{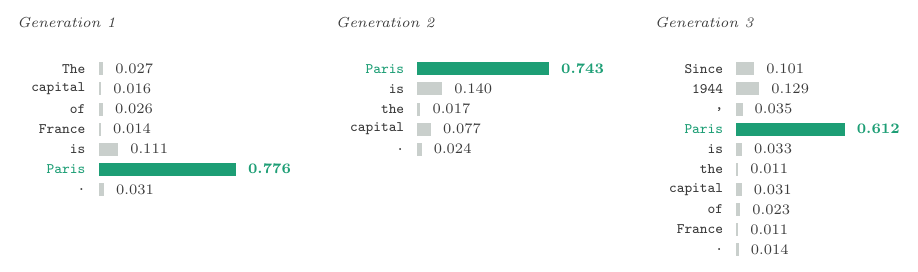}}
\caption{\textbf{Token weights concentrate on the answer span.} Weights $w_t$ from
Eq.~\ref{eq:te_enc} for three generations answering \emph{``What is the capital of
France?''}. The answer token carries the majority of the weight wherever it appears in the
sentence, while the tokens naming the question's subject receive almost none.}
\label{fig:token_weights}
\end{figure}

\subsection{Sensitivity to the Clustering Threshold}
\label{app:threshold}

SE and CAE require a binary matrix, so the continuous $\mS^{\mathrm{E}}$ must be
thresholded before the aggregator can be applied. We use $\tau = 0.5$ as a natural threshold.

Tab.~\ref{tab:threshold} reports $\mathrm{CAE}_{\mathrm{Enc}}$ as $\tau$ is varied.
Performance rises monotonically over the range tested, from $0.656$ at $\tau = 0.3$ to
$0.751$ at $\tau = 0.7$, and $\tau = 0.5$ is therefore not the best available setting:
$\tau = 0.7$ is $0.027$ higher. The sweep has not reached a maximum, so a still stricter
threshold might do better again.

We report the sweep rather than adopt its best point. The framework of Sec.~\ref{sec:method}
is trained entirely outside the uncertainty pipeline, with no UQ supervision and on data
disjoint from every evaluation setting, and selecting $\tau$ against the evaluation results
would introduce exactly the dependence the rest of the design avoids. The numbers in
Tab.~\ref{tab:main} are therefore obtained with the principled value, and the gains
reported there understate what a tuned threshold would give. %

\begin{table}[h!]
\centering
\caption{\textbf{Sensitivity to the clustering threshold.} AUROC and ECE of
$\mathrm{CAE}_{\mathrm{Enc}}$ over the 18 model--dataset combinations, as the threshold
$\tau$ applied to $\mS^{\mathrm{E}}$ is varied. The value used throughout the paper is
shaded.}
\label{tab:threshold}
\begin{tabular}{lcc}
\toprule
& AUROC & ECE \\
\midrule
$\tau = 0.3$ & .656 & .139 \\
$\tau = 0.4$ & .693 & .138 \\
\ours $\tau = 0.5$ & .724 & .140 \\
$\tau = 0.6$ & .743 & .137 \\
$\tau = 0.7$ & .751 & .141 \\
\bottomrule
\end{tabular}
\end{table}

\section{Training the operator}
\label{supp:training_details}
\subsection{Training Hyperparameters}

We fine-tune a \texttt{sentence-t5-large} encoder using triplet loss (see Eq.~\ref{eq:triplet_loss}) with
cosine distance and a margin of $0.2$, training with a batch size of $32$ and
a learning rate of $2\times10^{-5}$ under a linear warmup/decay schedule. We
train for up to $30$ epochs, with early stopping after $3$ epochs without
improvement in validation triplet accuracy. 

\subsection{Construction of the training set}
\label{app:training_data}

\paragraph{Generation and labelling.}
We use the complete \textbf{NQ-OPEN} val split ($3{,}160$ questions). For each we
sample $N = 10$ responses from \texttt{Llama-3.1-8B-Instruct} with temperature $1.0$, top-$p$ $0.9$
and top-$k$ $50$, matching the sampling regime used at evaluation so that the operator is
trained on the kind of text it will later compare. Each generation is labelled correct or
incorrect against the reference by the procedure of App.~\ref{supp:llm_judge}, partitioning
the $10$ generations of each question into a correct subset $\mathcal{P}$ and an incorrect
subset $\mathcal{N}$. No model or dataset used in Sec.~\ref{sec:experiments} appears at any
point in this pipeline.

\paragraph{Triplets.}
A question is retained only when $|\mathcal{P}| \geq 2$ and $|\mathcal{N}| \geq 1$, leaving
$1{,}720$ of the $3{,}160$. For each retained question a triplet is drawn uniformly as
\begin{equation*}
  \vy^{a} \sim \mathcal{P}, \qquad
  \vy^{+} \sim \mathcal{P} \setminus \{\vy^{a}\}, \qquad
  \vy^{-} \sim \mathcal{N},
\end{equation*}
with all three sharing the same question, which is prepended to each before encoding. Both
the anchor and the positive come from $\mathcal{P}$: two answers can be incorrect without
being incorrect in the same way, so a pair drawn from $\mathcal{N}$ would not reliably be
equivalent, and using such pairs as positives would teach the encoder to pull distinct
errors together. Triplets are resampled at every epoch, so each question contributes a
different pair each time it is visited.

\section{LLM as a Judge}
\label{supp:llm_judge}
We rely on an LLM-as-a-judge procedure at several stages of our pipeline. This section clarifies exactly how and where it is used.

\paragraph{Judging protocol.}
Given a question, an answer generated by the model under study, and the ground-truth answer, we prompt a larger external LLM to decide whether the generated answer matches the ground truth. The judge is never asked whether the generated answer is correct in its own opinion. Its role is restricted to comparing the generated answer with the reference, which accounts for paraphrases, formatting differences, or partial verbosity that exact-match or token-overlap metrics would miss. This text-to-text comparison is a simple task for a large instruction-tuned model. We use this procedure to annotate the training set.

\paragraph{Assigning a correctness label to each question.}
Evaluating uncertainty quantification (UQ) methods requires a single binary label per question, and two strategies are common in the literature. In the first, a greedy answer is generated alongside the sampled answers used to compute the uncertainty measures, and this greedy answer is labelled with the judging protocol above. In the second, each of the $N=20$ sampled answers is labelled individually, and the question receives the majority label among them. Both strategies are reasonable. We adopt the greedy-answer strategy for our main results because it is the more direct of the two.

\paragraph{Separate judges for training and evaluation.}
To rule out any dependency between the labels used for training and those used for evaluation, we use a different judge for each split: Qwen2.5-32B-Instruct annotates the training set, and Llama-3.1-70B-Instruct annotates the test set. We have no specific reason to expect such a dependency, but using distinct judges removes it by construction.

\paragraph{Robustness to the labeling strategy.}
The performance of our method does not depend on the aggregation strategy. The main results table reports scores computed with greedy-answer labels, and App.~\ref{supp:majority_vote} compares results under both strategies, showing that the conclusions are consistent.
\section{Evaluation protocol}
\label{supp:eval_details}

\subsection{Models, datasets and splits}
\label{supp:eval_data}
We evaluate on six datasets, again covering both settings, each on a fixed
subset of an official split. The visual question answering benchmarks are
\textbf{ADVQA}~\citep{li2021adversarial} (last 2{,}200 questions of the
validation set), \textbf{OKVQA}~\citep{marino2019ok} (first 5{,}000 of the
validation set) and \textbf{VizWiz}~\citep{gurari2018vizwiz} (last 4{,}400 of
the training set); the textual ones are \textbf{HotpotQA}~\citep{yang2018hotpotqa}
(first 5{,}000 of the validation set), \textbf{TriviaQA}~\citep{joshi2017triviaqa}
(first 1{,}000 of the validation set) and
\textbf{WebQuestions}~\citep{berant-etal-2013-semantic} (first 2{,}000 of the
test set). Each model is evaluated on the three datasets of its modality, giving
the 18 model--dataset combinations reported throughout.

\subsection{Prompts and decoding}
\label{supp:eval_prompts}

For every question, we sample $N=20$ answers with nucleus sampling
(temperature $1.0$, top-$p=0.9$, top-$k=50$). Each model is prompted with the original
question of the dataset, preceded by a short instruction: ``Answer the question
about this image.'' for visual benchmarks and ``Answer the following
question.'' for textual ones. We deliberately impose no constraint on the length
or form of the answer, so that the sampled generations retain their natural
variability in phrasing and verbosity, which is precisely what the operator
must see through.

\subsection{Estimator hyperparameters}
\label{supp:estimator_hparams}
SE, CAE and KLE use the entailment model
\texttt{microsoft/deberta-v2-xlarge-mnli}~\citet{he2020deberta} as their operator. For KLE, we report the heat-kernel variant with
$t=0.3$ and the Mat\'ern variant with $\kappa=1.0$ and $\nu=1.0$, the
hyperparameters that gave the best results for this baseline. COS-int embeds
each generated answer using the last-layer hidden states of the generating model
itself, and COS-ext uses \texttt{sentence-t5-large}~\citet{ni2022sentence}, the pretrained sentence
encoder from which our operator is fine-tuned. COS-ext and $\mathrm{COS}_{\mathrm{Enc}}$ therefore
share the same architecture and initialization, and differ only in our
training.

\section{Limits of existing operators}
\label{supp:limit_op}

\subsection{The context trap (COS-ext)}
\label{supp:limit_op_context_trap}

To test whether a general-purpose encoder separates answers that differ only on the fact
that matters, we construct pairs that are near-identical in phrasing but factually
opposed. The pipeline, applied to the \textbf{NQ-OPEN} train split, is:

\begin{enumerate}
  \item \textbf{Generate.} For each query, take the model's greedy decoded answer together
  with the list of ground-truth reference strings.
  \item \textbf{Filter.} Retain the entry only if the greedy answer contains an exact
  substring match to at least one reference, so that every retained example is verifiably
  correct rather than partial or paraphrased.
  \item \textbf{Corrupt.} Prompt \texttt{Llama-3.1-8B-Instruct} with the question, the
  matched ground truth and the greedy answer, asking it to replace the correct fact with a
  plausible but different incorrect one, preserving sentence structure and style.
\end{enumerate}

\noindent This yields $1{,}000$ pairs that agree in topic, length, and phrasing and disagree
only on the fact determining correctness:

\begin{center}
\footnotesize
\begin{tabular}{{l|c}}
\toprule
\emph{Question} & ``Who wrote the book The Origin of Species?''\\
\emph{Greedy (correct)} & The Origin of Species was written by \textbf{Charles Darwin} and
published in 1859. \\
\emph{Rewritten (incorrect)} & The Origin of Species was written by \textbf{Alfred Wallace} and
published in 1859. \\
\bottomrule
\end{tabular}
\end{center}

\noindent On these pairs, the base encoder gives a mean cosine similarity of $0.91$: it
does not separate them. The same encoder after our training gives $0.45$.

\subsection{The autoregressive bias (COS-int)}
\label{supp:limit_op_suffix_bias}

Hidden-state representations are read at the final answer token, so they are shaped most
by what the sequence ends on. To test whether this displaces the fact that determines
correctness, we construct pairs that end identically and disagree earlier. The pipeline,
applied to the \textbf{NQ-OPEN} train split, is:

\begin{enumerate}
  \item \textbf{Sample and label.} For each query, take the $10$ sampled answers and their
  per-sample correctness labels, and split them into a correct and an incorrect subset.
  \item \textbf{Match on suffix.} Search for a correct/incorrect pair sharing their last
  four words, so that the two answers converge on an identical closing phrase while
  disagreeing on the fact stated earlier in the sentence.
  \item \textbf{Extract.} Teacher-force each answer as the continuation of the question in
  a single forward pass with \texttt{Llama-3.1-8B-Instruct}, and take the final-layer hidden state at the last answer token.
\end{enumerate}

\noindent Natural samples rarely end on the same short phrase while diverging in
correctness, so the criterion is restrictive: 491 queries yield such a pair.

\begin{center}
\begin{tabular}{l|c}
\toprule
\emph{Question} & ``Who played the Joker in The Dark Knight'' \\
\emph{Correct} & Heath Ledger \textbf{played the role in the film.} \\
\emph{Incorrect} & Jack Nicholson \textbf{played the role in the film.} \\
\bottomrule
\end{tabular}
\end{center}

\noindent The answers differ only in their opening span, which carries the entire answer,
and agree on everything the representation reads last. Cosine similarity between the two
hidden states is $0.869$ on average: the representation is dominated by the shared suffix
rather than by the content preceding it. Our operator, on the same pairs, gives $0.350$.

\subsection{The entailment trap (NLI cross-encoders)}
\label{supp:limit_op_entailement_trap}

Entailment is a directional relation, and two answers can convey the same fact while
differing in how much else they say. To test whether this fractures valid semantic
clusters, we construct pairs that are equivalent by construction but maximally apart in
verbosity. The pipeline, applied to the \textbf{NQ-OPEN} train split, is:

\begin{enumerate}
  \item \textbf{Sample and label.} For each query, take the $20$ sampled answers and their
  per-sample correctness labels.
  \item \textbf{Keep the correct ones.} Discard every answer not labelled correct, so that
  each remaining answer is independently verified as conveying the reference fact.
  \item \textbf{Pair the extremes.} Pair the shortest correct answer with the longest,
  giving one pair per query spanning the range of natural response verbosity.
\end{enumerate}

\noindent This yields 1{,}000 pairs in which a concise entity-only answer is matched
against a verbose paraphrase of the same fact.

\begin{center}
\begin{tabular}{l|c}
\toprule
\emph{Question} & ``Who wrote the book The Origin of Species?'' \\
\emph{Shortest} & \textbf{Charles Darwin}. \\
\emph{Longest}  & The Origin of Species was written by \textbf{Charles
Darwin}, who published it in 1859. \\
\bottomrule
\end{tabular}
\end{center}

\noindent Averaging the predicted entailment probability in both directions, the NLI
cross-encoder scores these pairs at $0.291$, although every pair is factually equivalent
by construction: the additional correct detail of the longer answer blocks entailment in
one direction, and the objective penalizes verbosity rather than tracking agreement. Our
operator gives $0.801$ on the same pairs.

\subsection{The distribution mismatch (OOD syntax)}
\label{supp:limit_op_ood_syntax}

Beyond architectural mismatches, supervised metric spaces suffer a domain shift: operators
trained on human-annotated corpora expect well-formed sentences, while QA references are
often bare entities. To quantify this, we compare each generated sentence against the
reference string it matches. The pipeline, applied to the \textbf{NQ-OPEN} train split, is:

\begin{enumerate}
  \item \textbf{Generate.} For each query, take the model's greedy decoded answer.
  \item \textbf{Retrieve the reference.} Take the dataset's raw ground-truth string, which
  is typically a single entity rather than a sentence.
  \item \textbf{Pair.} Keep the pair when the generated answer is labelled correct, so
  that every pair is equivalent by construction while differing in syntactic form.
\end{enumerate}

\noindent This yields 1{,}000 pairs of a complete sentence against a bare entity.

\begin{center}
\begin{tabular}{l|c}
\toprule
\emph{Question} & ``What is the capital of France?'' \\
\emph{Generated} & The capital of France is \textbf{Paris}. \\
\emph{Reference} & \textbf{Paris} \\
\bottomrule
\end{tabular}
\end{center}

\noindent Both conventional operators collapse on these pairs: the NLI cross-encoder
scores them at $0.265$ and the base sentence encoder at $0.362$, although each pair is a trivially
correct match. Both conflate syntactic completeness with semantic overlap, and so fail on
precisely the fragmented strings that QA datasets use as references. Our operator reaches
$0.574$ on the same pairs.

\section{Complete results}
\label{supp:additional_results}

\subsection{Per-cell results}
\label{supp:per_cell}

Tab.~\ref{tab:percell_text} and Tab.~\ref{tab:percell_vqa} report every estimator's
per-model AUROC and ECE with bootstrapped 95\% confidence intervals, for the textual and
visual benchmarks, respectively, underlying the averages of Tab.~\ref{tab:main}.

\begin{table*}[h!]
\caption{\textbf{Per-cell results on the textual benchmarks.} AUROC ($\uparrow$) and ECE
($\downarrow$) with bootstrapped 95\% confidence intervals, for each estimator in its
published form (\emph{base}) and with our operator substituted (\emph{shaded}), reported
for each of the three text-only models.}
\label{tab:percell_text}
\centering
\resizebox{\textwidth}{!}{%
\begin{tabular}{l cc cc cc}
\toprule
& \multicolumn{2}{c}{\texttt{Qwen2.5-7B}} & \multicolumn{2}{c}{\texttt{Phi-3.5-mini}}
& \multicolumn{2}{c}{\texttt{Mistral-7B}} \\
Method & AUROC & ECE & AUROC & ECE & AUROC & ECE \\
\midrule
\multicolumn{7}{l}{{\textbf{HotpotQA}}} \\
CAE & .683 {\tiny ±.070} & .234 {\tiny ±.050} & .651 {\tiny ±.021} & .259 {\tiny ±.014} & .685 {\tiny ±.016} & .212 {\tiny ±.012} \\
\ours $\mathrm{CAE}_{\mathrm{Enc}}$ & \textbf{.724} {\tiny ±.067} & \textbf{.216} {\tiny ±.051} & \textbf{.722} {\tiny ±.019} & \textbf{.235} {\tiny ±.014} & \textbf{.719} {\tiny ±.015} & \textbf{.193} {\tiny ±.012} \\
SE & .684 {\tiny ±.067} & .234 {\tiny ±.046} & .653 {\tiny ±.020} & .258 {\tiny ±.014} & .686 {\tiny ±.015} & .212 {\tiny ±.013} \\
\ours $\mathrm{SE}_{\mathrm{Enc}}$ & \textbf{.726} {\tiny ±.064} & \textbf{.215} {\tiny ±.050} & \textbf{.726} {\tiny ±.021} & \textbf{.235} {\tiny ±.014} & \textbf{.722} {\tiny ±.016} & \textbf{.193} {\tiny ±.012} \\
KLE-Heat & .709 {\tiny ±.066} & \textbf{.207} {\tiny ±.050} & .714 {\tiny ±.019} & \textbf{.231} {\tiny ±.014} & .733 {\tiny ±.015} & \textbf{.186} {\tiny ±.013} \\
\ours $\mathrm{KLE\text{-}Heat}_{\mathrm{Enc}}$ & \textbf{.761} {\tiny ±.061} & .223 {\tiny ±.050} & \textbf{.762} {\tiny ±.019} & .240 {\tiny ±.015} & \textbf{.765} {\tiny ±.014} & .198 {\tiny ±.012} \\
KLE-Matern & .704 {\tiny ±.064} & \textbf{.225} {\tiny ±.049} & .703 {\tiny ±.020} & \textbf{.249} {\tiny ±.014} & .738 {\tiny ±.015} & \textbf{.203} {\tiny ±.012} \\
\ours $\mathrm{KLE\text{-}Matern}_{\mathrm{Enc}}$ & \textbf{.760} {\tiny ±.060} & .226 {\tiny ±.046} & \textbf{.762} {\tiny ±.019} & .243 {\tiny ±.014} & \textbf{.766} {\tiny ±.014} & .201 {\tiny ±.012} \\
COS-int & .752 {\tiny ±.058} & \textbf{.221} {\tiny ±.050} & .688 {\tiny ±.034} & .253 {\tiny ±.025} & .719 {\tiny ±.014} & .210 {\tiny ±.012} \\
COS-ext & .709 {\tiny ±.063} & .229 {\tiny ±.050} & .702 {\tiny ±.020} & \textbf{.252} {\tiny ±.014} & .709 {\tiny ±.016} & \textbf{.208} {\tiny ±.012} \\
\ours $\mathrm{COS}_{\mathrm{Enc}}$ & \textbf{.761} {\tiny ±.060} & .228 {\tiny ±.051} & \textbf{.760} {\tiny ±.020} & .246 {\tiny ±.014} & \textbf{.775} {\tiny ±.013} & \textbf{.203} {\tiny ±.012} \\
TE & \textbf{.785} {\tiny ±.057} & .219 {\tiny ±.049} & \textbf{.704} {\tiny ±.020} & .245 {\tiny ±.014} & \textbf{.725} {\tiny ±.016} & .200 {\tiny ±.013} \\
\ours $\mathrm{TE}_{\mathrm{Enc}}$ & .730 {\tiny ±.062} & \textbf{.215} {\tiny ±.049} & .688 {\tiny ±.020} & \textbf{.240} {\tiny ±.014} & .679 {\tiny ±.016} & \textbf{.194} {\tiny ±.012} \\
\midrule

\multicolumn{7}{l}{{\textbf{TriviaQA}}} \\
CAE & .677 {\tiny ±.036} & .165 {\tiny ±.032} & .462 {\tiny ±.034} & .232 {\tiny ±.031} & .611 {\tiny ±.038} & .240 {\tiny ±.027} \\
\ours $\mathrm{CAE}_{\mathrm{Enc}}$ & \textbf{.847} {\tiny ±.025} & \textbf{.126} {\tiny ±.024} & \textbf{.841} {\tiny ±.025} & \textbf{.087} {\tiny ±.022} & \textbf{.832} {\tiny ±.030} & \textbf{.191} {\tiny ±.027} \\
SE & .672 {\tiny ±.033} & .164 {\tiny ±.029} & .458 {\tiny ±.038} & .232 {\tiny ±.030} & .602 {\tiny ±.039} & .240 {\tiny ±.027} \\
\ours $\mathrm{SE}_{\mathrm{Enc}}$ & \textbf{.848} {\tiny ±.025} & \textbf{.121} {\tiny ±.023} & \textbf{.843} {\tiny ±.026} & \textbf{.088} {\tiny ±.022} & \textbf{.831} {\tiny ±.030} & \textbf{.190} {\tiny ±.028} \\
KLE-Heat & .789 {\tiny ±.029} & \textbf{.108} {\tiny ±.026} & .644 {\tiny ±.038} & \textbf{.084} {\tiny ±.028} & .808 {\tiny ±.033} & \textbf{.189} {\tiny ±.026} \\
\ours $\mathrm{KLE\text{-}Heat}_{\mathrm{Enc}}$ & \textbf{.860} {\tiny ±.023} & .122 {\tiny ±.024} & \textbf{.862} {\tiny ±.022} & .100 {\tiny ±.022} & \textbf{.845} {\tiny ±.027} & .196 {\tiny ±.026} \\
KLE-Matern & .777 {\tiny ±.030} & \textbf{.114} {\tiny ±.026} & .615 {\tiny ±.035} & \textbf{.137} {\tiny ±.029} & .795 {\tiny ±.035} & \textbf{.197} {\tiny ±.028} \\
\ours $\mathrm{KLE\text{-}Matern}_{\mathrm{Enc}}$ & \textbf{.859} {\tiny ±.022} & .124 {\tiny ±.025} & \textbf{.861} {\tiny ±.023} & .101 {\tiny ±.022} & \textbf{.844} {\tiny ±.027} & .199 {\tiny ±.027} \\
COS-int & .736 {\tiny ±.031} & .121 {\tiny ±.027} & .580 {\tiny ±.034} & .157 {\tiny ±.032} & .658 {\tiny ±.022} & .139 {\tiny ±.027} \\
COS-ext & \textbf{.790} {\tiny ±.029} & \textbf{.117} {\tiny ±.026} & \textbf{.808} {\tiny ±.028} & \textbf{.074} {\tiny ±.025} & \textbf{.756 }{\tiny ±.034} & \textbf{.197} {\tiny ±.027} \\
\ours $\mathrm{COS}_{\mathrm{Enc}}$ & \textbf{.855} {\tiny ±.023} & .118 {\tiny ±.026} & \textbf{.863} {\tiny ±.022} & .104 {\tiny ±.024} & \textbf{.838} {\tiny ±.028} & .200 {\tiny ±.027} \\
TE & .720 {\tiny ±.033} & .114 {\tiny ±.028} & .659 {\tiny ±.034} & .095 {\tiny ±.028} & .628 {\tiny ±.039} & .215 {\tiny ±.027} \\
\ours $\mathrm{TE}_{\mathrm{Enc}}$ & \textbf{.800} {\tiny ±.028} & \textbf{.103} {\tiny ±.026} & \textbf{.774} {\tiny ±.028} & \textbf{.061} {\tiny ±.024} & \textbf{.734} {\tiny ±.034} & \textbf{.193} {\tiny ±.030} \\
\midrule
\multicolumn{7}{l}{{\textbf{WebQuestions}}} \\
CAE & .558 {\tiny ±.025} & .137 {\tiny ±.022} & .438 {\tiny ±.025} & .206 {\tiny ±.021} & .474 {\tiny ±.026} & .186 {\tiny ±.022} \\
\ours $\mathrm{CAE}_{\mathrm{Enc}}$ & \textbf{.737} {\tiny ±.021} & \textbf{.056} {\tiny ±.016} & \textbf{.729} {\tiny ±.021} & \textbf{.067} {\tiny ±.019} & \textbf{.681} {\tiny ±.023} & \textbf{.042} {\tiny ±.019} \\
SE & .557 {\tiny ±.025} & .136 {\tiny ±.022} & .436 {\tiny ±.025} & .202 {\tiny ±.021} & .470 {\tiny ±.025} & .186 {\tiny ±.022} \\
\ours $\mathrm{SE}_{\mathrm{Enc}}$ & \textbf{.736} {\tiny ±.022} & \textbf{.058} {\tiny ±.017} & \textbf{.733} {\tiny ±.021} & \textbf{.067} {\tiny ±.020} & \textbf{.689} {\tiny ±.025} & \textbf{.047} {\tiny ±.019} \\
KLE-Heat & .597 {\tiny ±.025} & .099 {\tiny ±.021} & .530 {\tiny ±.027} & .148 {\tiny ±.021} & .538 {\tiny ±.024} & .089 {\tiny ±.021} \\
\ours $\mathrm{KLE\text{-}Heat}_{\mathrm{Enc}}$ & \textbf{.755} {\tiny ±.021} & \textbf{.035} {\tiny ±.017} & \textbf{.751} {\tiny ±.019} & \textbf{.058} {\tiny ±.018} & \textbf{.712} {\tiny ±.022} & \textbf{.041} {\tiny ±.016} \\
KLE-Matern & .598 {\tiny ±.025} & .106 {\tiny ±.019} & .513 {\tiny ±.025} & .162 {\tiny ±.021} & .534 {\tiny ±.024} & .147 {\tiny ±.022} \\
\ours $\mathrm{KLE\text{-}Matern}_{\mathrm{Enc}}$ & \textbf{.755} {\tiny ±.020} & \textbf{.033} {\tiny ±.018} & \textbf{.751} {\tiny ±.021} & \textbf{.058} {\tiny ±.021} & \textbf{.711} {\tiny ±.022} & \textbf{.041} {\tiny ±.016} \\
COS-int & .610 {\tiny ±.025} & .164 {\tiny ±.022} & .546 {\tiny ±.025} & .202 {\tiny ±.022} & .651 {\tiny ±.023} & .066 {\tiny ±.019} \\
COS-ext & .565 {\tiny ±.025} & .125 {\tiny ±.020} & .603 {\tiny ±.025} & .087 {\tiny ±.018} & .493 {\tiny ±.024} & .193 {\tiny ±.022} \\
\ours $\mathrm{COS}_{\mathrm{Enc}}$ & \textbf{.753} {\tiny ±.019} & \textbf{.028} {\tiny ±.016} & \textbf{.757} {\tiny ±.020} & \textbf{.055} {\tiny ±.020} & \textbf{.713} {\tiny ±.021} & \textbf{.046} {\tiny ±.017} \\
TE & .641 {\tiny ±.024} & .094 {\tiny ±.021} & .615 {\tiny ±.023} & .098 {\tiny ±.019} & .511 {\tiny ±.025} & .171 {\tiny ±.021} \\
\ours $\mathrm{TE}_{\mathrm{Enc}}$ & \textbf{.694} {\tiny ±.023} & \textbf{.051} {\tiny ±.019} & \textbf{.680} {\tiny ±.022} & \textbf{.075} {\tiny ±.020} & \textbf{.592} {\tiny ±.026} & \textbf{.097} {\tiny ±.021} \\
\bottomrule
\end{tabular}}
\end{table*}

\subsection{Question-conditioning ablation}
\label{supp:question_ablation}
The operator of Sec.~\ref{sec:method} reads the question alongside each generation, which
desiderata \textbf{(D2)} requires. To isolate that choice, we repeat the entire procedure without it: the encoder is
trained on answer strings alone, and evaluated the same way, so the comparison removes the
question from both stages rather than only from inference. Everything else is unchanged.

Tab.~\ref{tab:question} reports the result for $\mathrm{COS}_{\mathrm{Enc}}$. Mean AUROC
falls from $0.759$ to $0.742$. The gap is smaller than the gain over the conventional
operators, and we do not claim conditioning as the main source of the improvement, but it
is present in almost every one of the 18 model--dataset combinations, ranging from $+0.006$ to
$+0.026$. Its consistency across settings that share no model, no dataset and no modality
is what makes it interpretable as an effect of the conditioning rather than of any
particular evaluation. 

\begin{table}[h!]
\label{question_ablation}
\centering
\caption{\textbf{Conditioning on the question helps in every setting.} AUROC of
$\mathrm{COS}_{\mathrm{Enc}}$ with and without the question, by dataset (top) and by model
(bottom). In the answer-only variant, the question is removed from both training and
evaluation.}
\label{tab:question}
\resizebox{0.9\textwidth}{!}{
\begin{tabular}{lcccccc|c}
\toprule
& \textbf{ADVQA} & \textbf{OKVQA} & \textbf{VizWiz} & \textbf{HotpotQA} & \textbf{TriviaQA} & \textbf{WebQuestions} & Mean \\
\midrule
\ours with question & \textbf{.694} & \textbf{.758} & \textbf{.741} & \textbf{.765} & \textbf{.852} & \textbf{.741} & \textbf{.759} \\
answer only   & .668 & .735 & .730 & .759 & .842 & .719 & .742 \\
\midrule
& \texttt{idefics2-8B} & \texttt{Qwen2.5-VL-7B} & \texttt{llava-1.5-7B} & \texttt{Qwen2.5-7B} & \texttt{Phi-3.5-mini} & \texttt{Mistral-7B} & Mean \\
\midrule
 \ours with question & \textbf{.730} & \textbf{.740} & \textbf{.723} & \textbf{.790} & \textbf{.793} & \textbf{.775} & \textbf{.759} \\
answer only   & .711 & .725 & .697 & .776 & .782 & .761 & .742 \\
\bottomrule
\end{tabular}
}
\end{table}

\subsection{Greedy vs.\ majority-vote labels}
\label{supp:majority_vote}
As described in App.~\ref{supp:llm_judge}, each question can be assigned a correctness label in
two ways: by judging a single greedy answer, which is what we use in the main results, or by taking
the majority label over the $N=20$ sampled answers. Tab.~\ref{tab:greedy_vs_majority} reports the
AUROC of every estimator under both strategies, together with the gain our variant brings over its
base counterpart. Absolute scores are slightly higher with majority-vote labels for all methods,
but the conclusions do not change. In every block, our variant outperforms its base counterpart
under both strategies; the ranking within each block is identical, and the gains are of similar
magnitude, slightly larger under majority vote.

\begin{table}[h]
\caption{\textbf{AUROC under greedy and majority-vote correctness labels.}
Scores are averaged over the 18 model--dataset combinations. $\Delta$ is the AUROC gain of our
variant over its base counterpart (over COS-ext for the COS block). The last column indicates
whether the ranking within the block is the same under both labelling strategies.
Best value per block in bold.}
\label{tab:greedy_vs_majority}
\centering
\resizebox{0.6\textwidth}{!}{
\begin{tabular}{l cc cc c}
\toprule
& \multicolumn{2}{c}{Greedy} & \multicolumn{2}{c}{Majority vote} & \\
\cmidrule(lr){2-3} \cmidrule(lr){4-5}
Method & AUROC & $\Delta$ & AUROC & $\Delta$ & Order preserved \\
\midrule
CAE & .612 & & .629 & & \\
\ours $\mathrm{CAE}_{\mathrm{Enc}}$ & \textbf{.724} & +.112 & \textbf{.759} & +.130 & \multirow{-2}{*}{\checkmark} \\
\midrule
SE & .610 & & .626 & & \\
\ours $\mathrm{SE}_{\mathrm{Enc}}$ & \textbf{.726} & +.116 & \textbf{.761} & +.135 & \multirow{-2}{*}{\checkmark} \\
\midrule
KLE-Heat & .677 & & .706 & & \\
\ours $\mathrm{KLE\text{-}Heat}_{\mathrm{Enc}}$ & \textbf{.756} & +.079 & \textbf{.794} & +.088 & \multirow{-2}{*}{\checkmark} \\
\midrule
KLE-Matern & .682 & & .713 & & \\
\ours $\mathrm{KLE\text{-}Matern}_{\mathrm{Enc}}$ & \textbf{.756} & +.074 & \textbf{.794} & +.081 & \multirow{-2}{*}{\checkmark} \\
\midrule
COS-int & .635 & & .653 & & \\
COS-ext & .679 & & .704 & & \\
\ours $\mathrm{COS}_{\mathrm{Enc}}$ & \textbf{.759} & +.080 & \textbf{.797} & +.093 & \multirow{-3}{*}{\checkmark} \\
\midrule
TE & .640 & & .645 & & \\
\ours $\mathrm{TE}_{\mathrm{Enc}}$ & \textbf{.678} & +.038 & \textbf{.690} & +.045 & \multirow{-2}{*}{\checkmark} \\
\bottomrule
\end{tabular}
}
\end{table}

\begin{table*}[t!]
\caption{\textbf{Per-cell results on the visual benchmarks.} AUROC ($\uparrow$) and ECE
($\downarrow$) with bootstrapped 95\% confidence intervals, for each estimator in its
published form (\emph{base}) and with our operator substituted (\emph{shaded}), reported
for each of the three vision-language models.}
\label{tab:percell_vqa}
\centering
\resizebox{\textwidth}{!}{%
\begin{tabular}{l cc cc cc}
\toprule
& \multicolumn{2}{c}{\texttt{idefics2-8B}} & \multicolumn{2}{c}{\texttt{Qwen2.5-VL-7B}}
& \multicolumn{2}{c}{\texttt{llava-1.5-7B}} \\
Method & AUROC & ECE & AUROC & ECE & AUROC & ECE \\
\midrule
\multicolumn{7}{l}{{\textbf{ADVQA}}} \\
CAE & .644 {\tiny ±.021} & .115 {\tiny ±.018} & .631 {\tiny ±.024} & .127 {\tiny ±.020} & .635 {\tiny ±.022} & .107 {\tiny ±.019} \\
\ours $\mathrm{CAE}_{\mathrm{Enc}}$ & \textbf{.649} {\tiny ±.022} & \textbf{.085} {\tiny ±.019} & \textbf{.698} {\tiny ±.021} & \textbf{.059} {\tiny ±.019} & \textbf{.642} {\tiny ±.022} & \textbf{.099} {\tiny ±.020} \\
SE & .646 {\tiny ±.022} & .111 {\tiny ±.019} & .629 {\tiny ±.025} & .129 {\tiny ±.020} & .636 {\tiny ±.022} & .103 {\tiny ±.018} \\
\ours $\mathrm{SE}_{\mathrm{Enc}}$ & \textbf{.652} {\tiny ±.023} & \textbf{.083} {\tiny ±.020} & \textbf{.702} {\tiny ±.022} & \textbf{.060} {\tiny ±.019} & \textbf{.645} {\tiny ±.023} & \textbf{.098} {\tiny ±.020} \\
KLE-Heat & .670 {\tiny ±.023} & .093 {\tiny ±.021} & .675 {\tiny ±.022} & .074 {\tiny ±.019} & \textbf{.679} {\tiny ±.022} & \textbf{.073} {\tiny ±.018} \\
\ours $\mathrm{KLE\text{-}Heat}_{\mathrm{Enc}}$ & \textbf{.678} {\tiny ±.020} & \textbf{.071} {\tiny ±.018} & \textbf{.712} {\tiny ±.020} & \textbf{.073} {\tiny ±.019} & .659 {\tiny ±.021} & .088 {\tiny ±.018} \\
KLE-Matern & \textbf{.686} {\tiny ±.021} & .073 {\tiny ±.017} & .690 {\tiny ±.021} & .070 {\tiny ±.017} & \textbf{.688} {\tiny ±.020} & \textbf{.067} {\tiny ±.017} \\
\ours $\mathrm{KLE\text{-}Matern}_{\mathrm{Enc}}$ & .679 {\tiny ±.022} & \textbf{.070} {\tiny ±.017} & \textbf{.712} {\tiny ±.021} & \textbf{.070} {\tiny ±.018} & .660 {\tiny ±.023} & .081 {\tiny ±.020} \\
COS-int & .571 {\tiny ±.024} & .121 {\tiny ±.021} & .582 {\tiny ±.024} & .131 {\tiny ±.020} & .579 {\tiny ±.023} & .101 {\tiny ±.020} \\
COS-ext & .641 {\tiny ±.022} & .094 {\tiny ±.018} & .684 {\tiny ±.022} & .087 {\tiny ±.020} & .629 {\tiny ±.024} & .101 {\tiny ±.020} \\
\ours $\mathrm{COS}_{\mathrm{Enc}}$ & \textbf{.694} {\tiny ±.022} & \textbf{.071} {\tiny ±.020} & \textbf{.707} {\tiny ±.021} & \textbf{.075} {\tiny ±.018} & \textbf{.682} {\tiny ±.022} & \textbf{.071} {\tiny ±.018} \\
TE & .601 {\tiny ±.025} & .103 {\tiny ±.020} & .612 {\tiny ±.023} & .115 {\tiny ±.020} & .611 {\tiny ±.023} & \textbf{.080} {\tiny ±.021} \\
\ours $\mathrm{TE}_{\mathrm{Enc}}$ & \textbf{.636} {\tiny ±.022} & \textbf{.082} {\tiny ±.019} & \textbf{.635} {\tiny ±.023} & \textbf{.074} {\tiny ±.020} & \textbf{.637} {\tiny ±.024} & \textbf{.080} {\tiny ±.020} \\
\midrule
\multicolumn{7}{l}{{\textbf{OKVQA}}} \\
CAE & .678 {\tiny ±.016} & .240 {\tiny ±.012} & .597 {\tiny ±.017} & .226 {\tiny ±.013} & .654 {\tiny ±.016} & .220 {\tiny ±.012} \\
\ours $\mathrm{CAE}_{\mathrm{Enc}}$ & \textbf{.724} {\tiny ±.015} & \textbf{.210} {\tiny ±.012} & \textbf{.749} {\tiny ±.015} & \textbf{.179} {\tiny ±.013} & \textbf{.725} {\tiny ±.015} & \textbf{.192} {\tiny ±.013} \\
SE & .677 {\tiny ±.016} & .239 {\tiny ±.013} & .598 {\tiny ±.017} & .223 {\tiny ±.013} & .652 {\tiny ±.017} & .221 {\tiny ±.012} \\
\ours $\mathrm{SE}_{\mathrm{Enc}}$ & \textbf{.723} {\tiny ±.016} & \textbf{.210} {\tiny ±.012} & \textbf{.749} {\tiny ±.015} & \textbf{.179} {\tiny ±.012} & \textbf{.727} {\tiny ±.016} & \textbf{.191} {\tiny ±.012} \\
KLE-Heat & .699 {\tiny ±.015} & \textbf{.206} {\tiny ±.012} & .652 {\tiny ±.016} & \textbf{.169} {\tiny ±.013} & .684 {\tiny ±.016} & \textbf{.186} {\tiny ±.013} \\
\ours $\mathrm{KLE\text{-}Heat}_{\mathrm{Enc}}$ & \textbf{.765} {\tiny ±.014} & .219 {\tiny ±.012} & \textbf{.764} {\tiny ±.014} & .189 {\tiny ±.013} & \textbf{.761} {\tiny ±.014} & .201 {\tiny ±.013} \\
KLE-Matern & .723 {\tiny ±.016} & \textbf{.220} {\tiny ±.012} & .654 {\tiny ±.016} & \textbf{.185} {\tiny ±.012} & .715 {\tiny ±.015} & \textbf{.201} {\tiny ±.013} \\
\ours $\mathrm{KLE\text{-}Matern}_{\mathrm{Enc}}$ & \textbf{.764} {\tiny ±.014} & .222 {\tiny ±.012} & \textbf{.763} {\tiny ±.014} & .190 {\tiny ±.013} & \textbf{.760} {\tiny ±.014} & .203 {\tiny ±.012} \\
COS-int & .592 {\tiny ±.017} & \textbf{.199} {\tiny ±.013} & .626 {\tiny ±.016} & .210 {\tiny ±.013} & .646 {\tiny ±.017} & \textbf{.201} {\tiny ±.012} \\
COS-ext & .674 {\tiny ±.017} & .224 {\tiny ±.012} & .677 {\tiny ±.016} & \textbf{.187} {\tiny ±.013} & .668 {\tiny ±.015} & .202 {\tiny ±.013} \\
\ours $\mathrm{COS}_{\mathrm{Enc}}$ & \textbf{.759} {\tiny ±.014} & .224 {\tiny ±.012} & \textbf{.763} {\tiny ±.014} & .193 {\tiny ±.013} & \textbf{.751} {\tiny ±.014} & .207 {\tiny ±.013} \\
TE & .613 {\tiny ±.017} & .223 {\tiny ±.012} & .566 {\tiny ±.017} & .186 {\tiny ±.012} & .602 {\tiny ±.016} & .206 {\tiny ±.012} \\
\ours $\mathrm{TE}_{\mathrm{Enc}}$ & \textbf{.660} {\tiny ±.016} & \textbf{.216} {\tiny ±.012} & \textbf{.610} {\tiny ±.016} & \textbf{.184} {\tiny ±.012} & \textbf{.653} {\tiny ±.016} & \textbf{.198} {\tiny ±.012} \\
\midrule
\multicolumn{7}{l}{{\textbf{VizWiz}}} \\
CAE & .641 {\tiny ±.017} & .219 {\tiny ±.013} & .649 {\tiny ±.017} & .235 {\tiny ±.013} & .653 {\tiny ±.015} & .142 {\tiny ±.013} \\
\ours $\mathrm{CAE}_{\mathrm{Enc}}$ & \textbf{.660} {\tiny ±.017} & \textbf{.183} {\tiny ±.013} & \textbf{.703} {\tiny ±.017} & \textbf{.194} {\tiny ±.013} & \textbf{.657} {\tiny ±.017} & \textbf{.101} {\tiny ±.015} \\
SE & .639 {\tiny ±.017} & .218 {\tiny ±.014} & .646 {\tiny ±.017} & .233 {\tiny ±.013} & .647 {\tiny ±.016} & .141 {\tiny ±.014} \\
\ours $\mathrm{SE}_{\mathrm{Enc}}$ & \textbf{.662} {\tiny ±.018} & \textbf{.182} {\tiny ±.013} & \textbf{.701} {\tiny ±.017} & \textbf{.193} {\tiny ±.014} & \textbf{.662} {\tiny ±.017} & \textbf{.101} {\tiny ±.014} \\
KLE-Heat & .693 {\tiny ±.017} & \textbf{.187} {\tiny ±.013} & .689 {\tiny ±.016} & \textbf{.191} {\tiny ±.014} & .689 {\tiny ±.015} & \textbf{.104} {\tiny ±.013} \\
\ours $\mathrm{KLE\text{-}Heat}_{\mathrm{Enc}}$ & \textbf{.728} {\tiny ±.016} & .193 {\tiny ±.013} & \textbf{.744} {\tiny ±.016} & .204 {\tiny ±.014} & \textbf{.725} {\tiny ±.015} & .112 {\tiny ±.015} \\
KLE-Matern & .711 {\tiny ±.016} & .202 {\tiny ±.013} & .700 {\tiny ±.016} & .211 {\tiny ±.014} & .725 {\tiny ±.015} & .122 {\tiny ±.014} \\
\ours $\mathrm{KLE\text{-}Matern}_{\mathrm{Enc}}$ & \textbf{.729} {\tiny ±.016} & \textbf{.197} {\tiny ±.013} & \textbf{.744} {\tiny ±.015} & \textbf{.207} {\tiny ±.013} & \textbf{.727} {\tiny ±.015} & \textbf{.116} {\tiny ±.015} \\
COS-int & .612 {\tiny ±.017} & \textbf{.183} {\tiny ±.013} & .665 {\tiny ±.018} & .215 {\tiny ±.013} & .645 {\tiny ±.016} & \textbf{.115} {\tiny ±.014} \\
COS-ext & .693 {\tiny ±.018} & .203 {\tiny ±.013} & .717 {\tiny ±.017} & .212 {\tiny ±.014} & .703 {\tiny ±.015} & .121 {\tiny ±.014} \\
\ours $\mathrm{COS}_{\mathrm{Enc}}$ & \textbf{.738} {\tiny ±.016} & .201 {\tiny ±.013} & \textbf{.749} {\tiny ±.016} & \textbf{.211} {\tiny ±.013} & \textbf{.736} {\tiny ±.014} & .121 {\tiny ±.015} \\
TE & .632 {\tiny ±.017} & .199 {\tiny ±.013} & .660 {\tiny ±.018} & .215 {\tiny ±.013} & .644 {\tiny ±.016} & .122 {\tiny ±.012} \\
\ours $\mathrm{TE}_{\mathrm{Enc}}$ & \textbf{.654} {\tiny ±.017} & \textbf{.194} {\tiny ±.014} & \textbf{.688} {\tiny ±.017} & \textbf{.204} {\tiny ±.014} & \textbf{.662} {\tiny ±.016} & \textbf{.116} {\tiny ±.014} \\
\bottomrule
\end{tabular}}
\end{table*}

\end{document}